\documentclass[11pt,twoside]{article}

\usepackage{epsfig,amsfonts,color}
\usepackage{amsmath}
\usepackage{amssymb, palatino, geometry,url}
\usepackage[noresetcount,lined,boxed]{algorithm2e} 

\usepackage[colorlinks=true,linkcolor=blue,citecolor=blue,urlcolor=blue]{hyperref}
\usepackage{amsmath}
\allowdisplaybreaks

\usepackage{fancyhdr}
\newcommand{\be}{\begin{equation}}
\newcommand{\ee}{\end{equation}}
\newcommand{\bea}{\begin{eqnarray}}
\newcommand{\eea}{\end{eqnarray}}

\newcommand{\bvec}{\left(\begin{array}{c}}
\newcommand{\evec}{\end{array}\right)}
\newcommand{\bsub}{\begin{subequations}}
\newcommand{\esub}{\end{subequations}}

\usepackage{lineno}
\usepackage{pdflscape}
\usepackage{subfigure}
\usepackage{algorithm2e}
\usepackage{tabularx}
\usepackage{booktabs}
\usepackage{siunitx}
\usepackage{multirow}
\usepackage{xurl}
\usepackage{caption}
\usepackage{xr}
\usepackage{nomencl}

\makenomenclature

\begin{document}

\title{On the Implementation of a Bayesian Optimization Framework for Interconnected Systems}

\author{Leonardo D. Gonz\'alez and Victor M. Zavala\thanks{Corresponding Author: victor.zavala@wisc.edu.}\\
 \\
  {\small Department of Chemical and Biological Engineering}\\
 {\small \;University of Wisconsin-Madison, 1415 Engineering Dr, Madison, WI 53706, USA}}
 \date{}
\maketitle

\begin{abstract}
Bayesian optimization \nomenclature{\textbf{BO}}{Bayesian Optimization} (BO) is an effective paradigm for the optimization of expensive-to-sample systems. Standard BO learns the performance of a system $f(x)$ by using a Gaussian Process (GP) model; this treats the system as a black-box and limits its ability to exploit available structural knowledge (e.g., physics and sparse interconnections in a complex system). Grey-box modeling, wherein the performance function is treated as a composition of known and unknown intermediate functions $f(x, y(x))$ (where $y(x)$ is a GP model) offers a solution to this limitation; however, generating an analytical probability density for $f$ from the Gaussian density of $y(x)$ is often an intractable problem (e.g., when $f$ is nonlinear). Previous work has handled this issue by using sampling techniques or by solving an auxiliary problem over an augmented space where the values of $y(x)$ are constrained by confidence intervals derived from the GP models; such solutions are computationally intensive. In this work, we provide a detailed implementation of a recently proposed grey-box BO paradigm, \nomenclature{\textbf{BOIS}}{Bayesian Optimization for Interconnected Systems algorithm} BOIS, that uses adaptive linearizations of $f$ to obtain analytical  expressions for the statistical moments of the composite function. We show that the BOIS approach enables the exploitation of structural knowledge, such as that arising in interconnected systems as well as systems that embed multiple GP models and combinations of physics and GP models. We benchmark the effectiveness of BOIS against standard BO and existing grey-box BO algorithms using a pair of case studies focused on chemical process optimization and design. Our results indicate that BOIS performs as well as or better than existing grey-box methods, while also being less computationally intensive.
\end{abstract}

{\bf Keywords}: Bayesian optimization, grey-box modeling, composite functions, linearization


\section{Introduction}

Optimization of complex systems such as chemical processes is often challenging due to incomplete physical knowledge and/or the need to simulate complex models or collect experimental data. This has motivated the development of black-box optimization strategies \cite{Conn:2009}; these methods use input/output data from the system to generate a surrogate model that is used to guide the search. In applications where system queries are often expensive and quantifying uncertainty in predicted performance is important, Bayesian optimization (BO) \cite{Mockus:2012} has emerged as one of the most effective black-box optimization paradigms.
\\

\textcolor{black}{The ability of BO to efficiently solve challenging problems has led to the development of a rich body of work detailing its history \cite{Garnett:2023, Brochu:2010}, benchmarking its performance against alternative optimizers \cite{Jones:1998, Snoek:2012, Wilson:2014, Greenhill:2020}, and exploring its use across several} disciplines, such as materials engineering \cite{Hase:2021}, aerospace engineering \cite{Lam:2018}, control tuning \cite{Sorourifar:2021}, and synthetic biology \cite{Radivojevic:2020}. BO is a flexible algorithm capable of accommodating both continuous and discrete design variables \cite{Brochu:2010}, handling problem constraints \cite{Priem:2019}, and identifying failure regions \cite{Chakrabarty:2023}. The most powerful feature of BO is, arguably, its ability to select sample points (experiments) effectively. Specifically, BO uses the collected input/output data to train a probabilistic surrogate model, typically a Gaussian process \nomenclature{\textbf{GP}}{Gaussian Process} (GP), that estimates not only the predicted system performance but also the uncertainty of the predictions. These estimates are used to construct an acquisition function \nomenclature{\textbf{AF}}{Acquisition Function} (AF), which serves as the decision-making mechanism of the algorithm, that assigns value to sample points based on both their information gain and expected performance. The AF can be constructed to place greater importance on sampling from regions with high predicted performance (exploitation) or high model uncertainty (exploration). This consideration of information value in addition to performance enables BO to efficiently sample from several distinct regions of the design space \cite{Shahriari:2016}.
\\

\begin{figure}[!htp]
	\centering
	\includegraphics[width=0.7\textwidth]{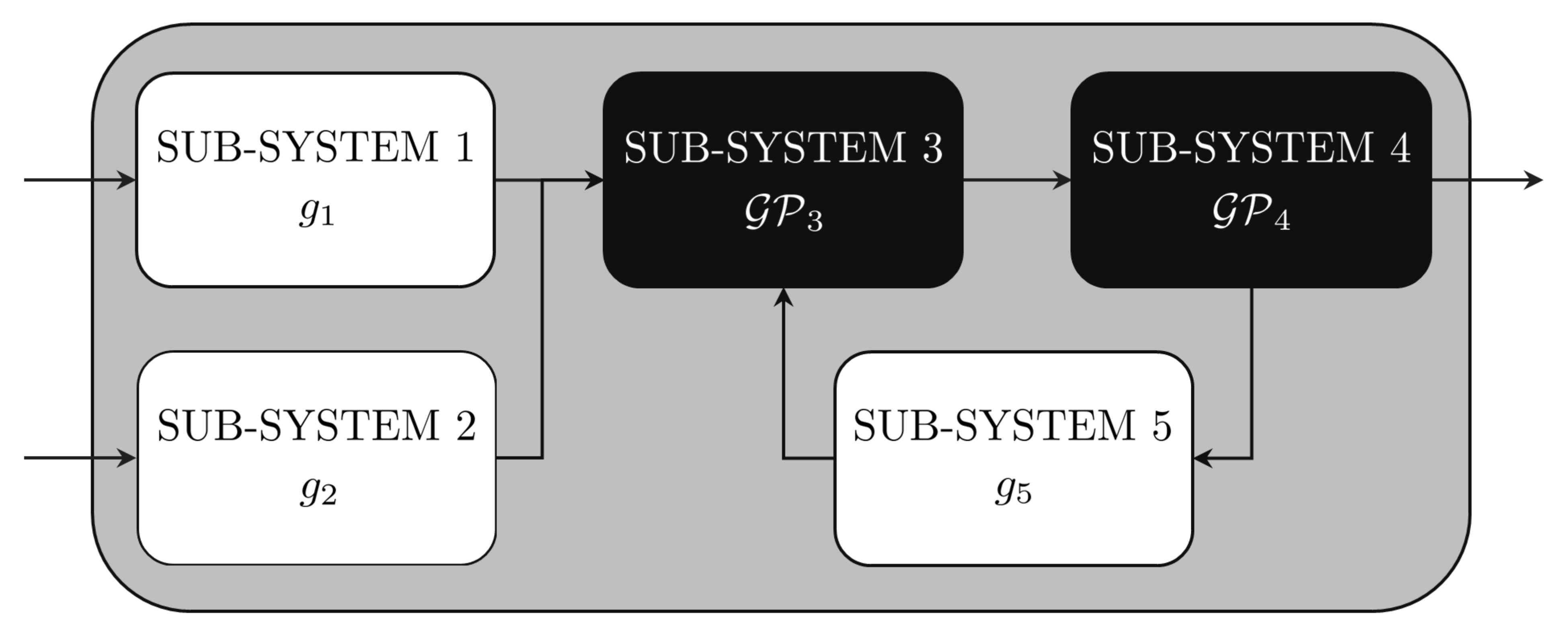}
	\caption{Grey-box systems often exhibit a known structure where the connectivity between different elements is understood. Not every component is always a black-box that requires a surrogate model as a closed-form representation might be available for various components.}
	\label{fig:grey_box_system}
\end{figure}

While the black-box assumption makes BO highly flexible (only an interface for providing inputs and collecting output data is needed), there is often some form of structural system knowledge available (e.g., physics or sparse interconnectivity of system components). For example, when dealing with a complex physical system (e.g., a chemical process), several components might be well-modeled and understood, while others might not. In other words, the system is actually a \textit{composition} of various white-box (i.e., an analytical representation is available) and black-box elements as shown in Figure \ref{fig:grey_box_system}. Furthermore, the fundamental principles governing the behavior of the black-box elements (e.g., conservation laws, equilibrium, value constraints) are, at least qualitatively, understood. Additionally, sparse connectivity, which provides information on how different components interact, is also often known. As a result, the system of interest is usually not truly a black-box but rather a grey-box that is partially observable with a known structure \cite{Sohlberg:2008}. Previous work done using several different optimization frameworks has demonstrated that exploiting this knowledge, as opposed to relying on a purely black-box strategy, can significantly improve the optimization search \cite{boukouvala:2017, Bajaj:2018, Beykal:2018}. 
\\

Various methods have been developed that allow for the consideration of grey-box models in BO. Most of these approaches involve the use of a low-fidelity approximation of the system that is cheaper to evaluate. The simplified representation is fed to the algorithm, allowing it to learn coarse system trends and to identify potentially promising regions of the design space from the start. The approximation can be obtained through a variety of means (e.g., simplified physical models, empirical correlations, or lower-fidelity simulations) and can either be gradually refined \cite{Kandasamy:2017, Sorourifar:2023, Wu:2021} or remain unchanged as the algorithm progresses \cite{Lu:2021}. 
\\

The work in \cite{Astudillo:2019} has led to a push towards developing BO frameworks that represent a system as a {\em composite function}, $f(x, y(x))$, where $x$ are the system inputs, $f$ is a known scalar function, and $y$ is an unknown vector-valued function that describes the behavior of internal system components. The composite representation shifts the modeling task from estimating the performance function directly to estimating the values of $y$ which serve as inputs to $f(x, y(x))$. This can result in derivative information for $f$ becoming available, allowing for a clearer understanding of the effects of $x$ and $y$ on system performance \cite{Urenholt:2019}. Additionally, such a representation allows for explicit separation of the white-box and black-box sections of the system, which can enable a reduction in the dimensionality of the surrogate models and allows for the modeling task to be redistributed to a simpler set of intermediate functions when $f$ is complex \cite{Xu:2023}. This approach also lends itself to the inclusion of constraints, as these are often dependent on internal variables which can be captured by $y$ \cite{Paulson:2022, Lu:2023}. As a result, composite functions allow for a more complete representation of a system, especially in the context of engineering design. For example, in chemical process design the cost equations for equipment, material streams, and utilities are often known but the parameters that equations rely on (e.g., flow rates and heat duties) might be unknown. Furthermore, traditional unit operations (e.g., heat exchangers, distillation columns, compressors) have significantly better physics models available than those that tend to be more specialized units (e.g., bioreactors, non-equilibrium separators, solids-handling). 
\\

While setting up a composite function optimization problem might be intuitive, implementing this in a BO setting is not trivial. One of the main advantages of BO is the inclusion of the uncertainty estimates of the surrogate model, which allows for greater exploration of the design space when compared to a deterministic model \cite{boukouvala:2017}. However, when using a composite function, the GP model generated is of $y$ and not of $f$. Given that $f$ is the performance metric that needs to be optimized, it is necessary to propagate the predicted uncertainty from $y(x)$ to $f(x, y(x))$ (i.e., the density of $f$ or desired summarizing statistics must be determined). A Gaussian density for $y(x)$ is directly obtained from the GP surrogate; when $f$ is a linear model, we can make use of the closure of Gaussian random variables under linear operations to generate the density of $f(x, y(x))$ (which is also a Gaussian). When $f$ is nonlinear, however, an analytical form is not readily available and alternative methods must be used to obtain the density. This problem has traditionally been solved numerically using sampling methods such as Monte Carlo \cite{Astudillo:2019, Astudillo:2021, Balandat:2020, Paulson:2022}; however, this approach can quickly become computationally intensive. An alternative method proposed in \cite{Xu:2023} avoids the need to explicitly generate a probability density for $f$ by utilizing the so-called optimism-driven algorithm that solves an auxiliary problem that is defined over an augmented space, allowing for the optimization to be carried out with respect to $x$ and $y$. The trained GP models are used to construct a set of lower and upper confidence bound functions that are incorporated into the augmented problem as constraints. This specifies a range from which values for $y$ can be selected based on the performance and uncertainty estimates of the GPs. This approach allows the algorithm to eliminate the need for sampling; however, it increases the size and complexity of the optimization task. This can significantly increase the computational time required to find a solution, especially when $x,y$ are high-dimensional.
\\

\begin{figure}[!htp]
	\centering
	\includegraphics[width=0.75\textwidth]{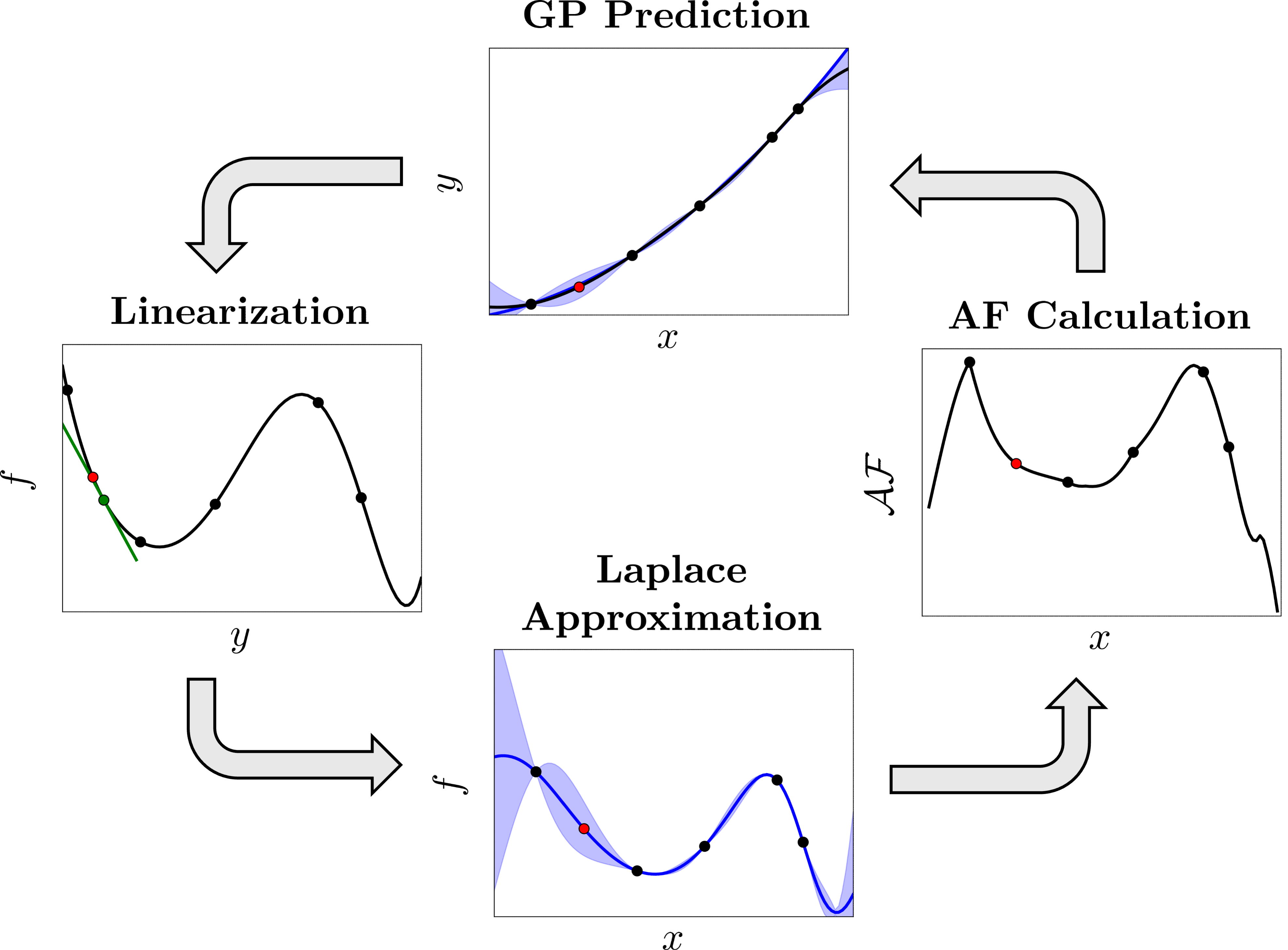}
	\caption{Illustration of the adaptive linearization scheme employed by BOIS. At a point $x$ of interest (red marker), a Gaussian process model estimates the value of intermediate $y$. A local Laplace approximation is then constructed by linearizing $f$ around a neighboring point (green marker). The summarizing statistics are passed into an acquisition function that determines the value of sampling at the selected point. This process is repeated until the optimum of the acquisition function is found.}
	\label{fig:adaptive_linearization}
\end{figure}

The increased functionality of composite functions coupled with the high computational intensity of existing methods motivates the need to develop more efficient paradigms for composite function BO. We recently proposed the Bayesian Optimization of Interconnected Systems (BOIS) framework, a new method that facilitates the use of composite functions via adaptive linearizations of $f(x, y(x))$ in the neighborhood of a $y(x)$ of interest (see Figure \ref{fig:adaptive_linearization}) \cite{Gonzalez:2023}. This allows for the construction of local Laplace approximations that can be used to generate closed-form analytical expressions for the mean and uncertainty/variance of $f$. In this work, we extend our analysis of the BOIS framework; specifically, by using a pair of complex case studies, we refine our implementation and provide further evidence of the performance and efficiency improvements this algorithm provides over standard BO as well as the composite function BO paradigms presented in \cite{Astudillo:2019} and \cite{Xu:2023}. Additionally, we implement new functionalities that allow us to handle feasibility considerations for the intermediate functions. \textcolor{black}{We also exploit the ability of this framework to build a nested function structure for $y$, wherein the dependencies of a given intermediate element on other elements in $y$ can be explicitly considered. This provides a significant degree of flexibility in the selection of the intermediate functions and facilitates the use of available white-box models.} This also allows us to reduce the number of surrogates that must be constructed to model $y$ and the dimensions of the corresponding input spaces of these models. As a result, we are able to develop black-box models that enable system-wide optimization in a more scalable and efficient manner than existing methods.

\newpage
\section{Standard Bayesian Optimization}

Consider the general optimization problem:
\begin{subequations} \label{eq:goal}
    \begin{gather}
	\min_x~~f(x)\\
	\textrm{s.t.}~~x\in X
    \end{gather}
\end{subequations}
where $f: X\to \mathbb{R}$ is a scalar performance function, $X\subseteq\mathbb{R}^{d_x}$ is the design space, and $x$ is a set of design inputs within $X$. Generally, solving this problem is made difficult by the fact that there is no analytical representation of the objective function. 

\begin{figure}[hbt!]
	\centering
	\includegraphics[width=0.75\textwidth]{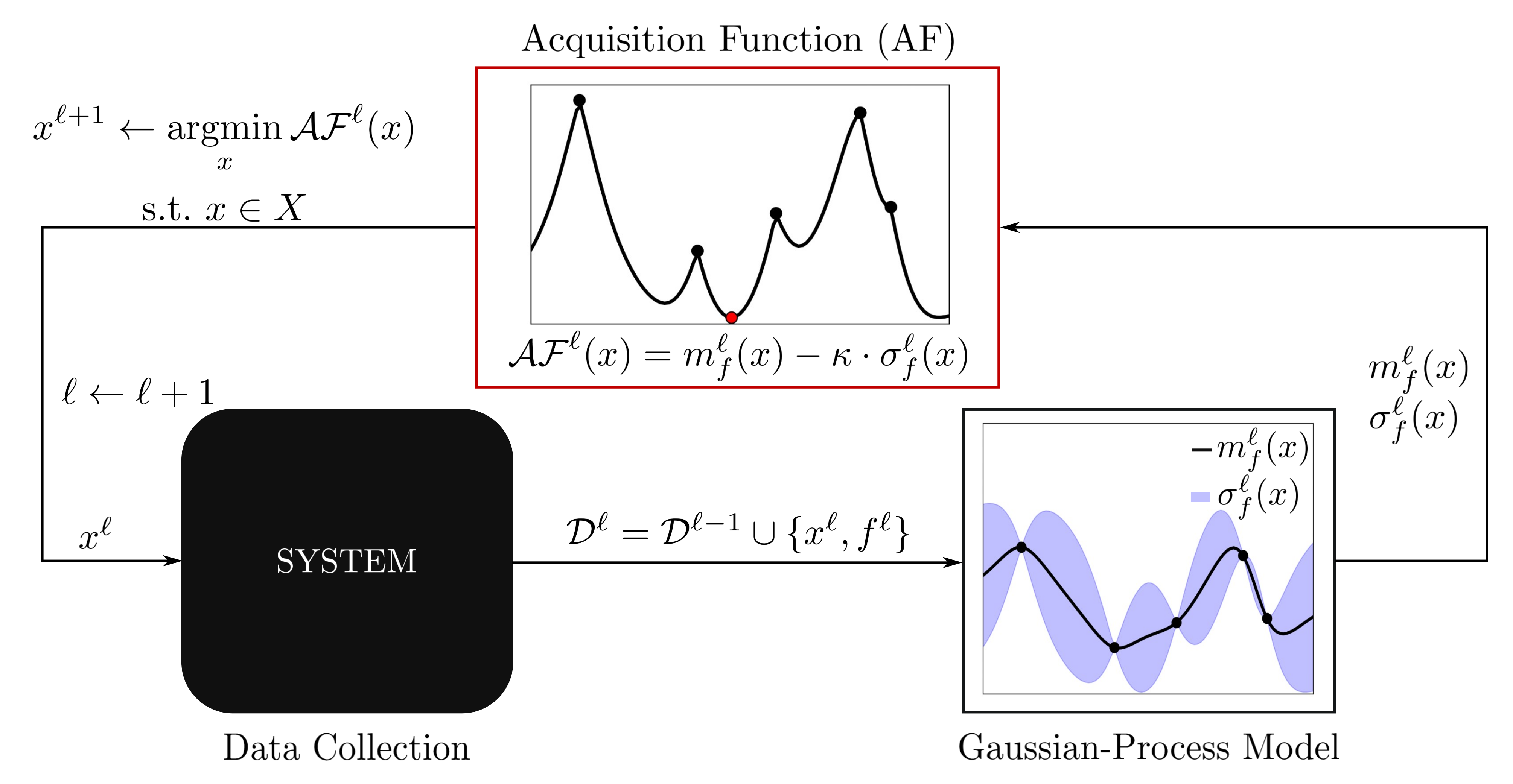}
	\caption{Workflow of the S-BO framework. Using a dataset $\mathcal{D}^{\ell}$, BO builds a GP surrogate model of the system. The mean and variance estimates calculated by the GP are passed into an acquisition function that is optimized to suggest a new sampling point $x^{\ell+1}$. The system is sampled at this point and the collected data is appended to the dataset to retrain the GP model.}
	\label{fig:S_BO}
\end{figure}

Standard Bayesian optimization \nomenclature{\textbf{S-BO}}{Standard Bayesian Optimization algorithm} (S-BO) solves the problem in \eqref{eq:goal} by generating a sequence of sample points (experiments) based on the results of prior observations \cite{Garnett:2023}. The value of sampling at a new point is measured by an acquisition function which considers not only its estimated performance but also its informational value. These are quantified via a GP surrogate model that is trained on the input/output data. The algorithm is initialized using a dataset of size ${\ell}$, $\mathcal{D}^{\ell} = \{x_\mathcal{K}, f_\mathcal{K}\}$, where $\mathcal{K}=\{1,...,\ell\}$, to train the GP. 
\\

The GP assumes that the output data follow a distribution of the form $f(x_\mathcal{K})\sim\mathcal{N}\left(\textbf{m}(x),\textbf{K}(x, x^\prime)\right)$ where $\textbf{m}(x)\in\mathbb{R}^{\ell}$ is the mean function and $\textbf{K}(x, x^\prime)\in\mathbb{R}^{\ell\times \ell}$ is the covariance matrix \cite{Rasmussen:2006}. While $\textbf{m}(x)$ is usually set equal to \textbf{0}, $\textbf{K}(x, x^\prime)$ is calculated using a kernel function, $k(x, x^\prime)$, such that $\textbf{K}_{ij}=k(x_i, x_j)$. In our work, we have opted to use the anisotropic M\'atern kernel \cite{Matern:1960} which is defined as:
\begin{equation}\label{eq:matern_kernel}
	k\left(x, x^{\prime}\right) = \frac{1}{\Gamma(\nu)2^{\nu-1}}\left(\sqrt{2\nu}d\left(x, x^{\prime}\right)\right)^\nu K_\nu\left(\sqrt{2\nu}d\left(x, x^{\prime}\right)\right)
\end{equation}
In \eqref{eq:matern_kernel}, $\nu$ denotes the smoothness of the generated function and is usually set to either 1.5 (function is once-differentiable) or 2.5 (function is twice-differentiable); $\Gamma$ is the gamma function and $K_\nu$ is a modified Bessel function. The function $d(x, x^{\prime}) = \sqrt{\left(x-x^{\prime}\right)^T\Theta^{-2}\left(x-x^{\prime}\right)}$ is a scaled Euclidean distance function. Here $\Theta\in\mathbb{R}^{d_x\times d_x}$ is a diagonal matrix whose entries are the kernel length scales, $\theta_1,...,\theta_{d_x}$, along each dimension of $x$. The length scales are also referred to as the kernel hyperparameters and their values are calculated by maximizing the log marginal likelihood function with respect to $\theta$ as  shown in \eqref {eq:LML_function}:
\begin{equation}\label{eq:LML_function}
	\theta^{\ast} \in \mathop{\textrm{argmax}}_{\theta}\; \log p(f_{\mathcal{K}}|x_{\mathcal{K}}, \theta) = \mathop{\textrm{argmax}}_{\theta} -\frac{1}{2}f_{\mathcal{K}}^T\textbf{K}^{-1}f_{\mathcal{K}}-\frac{1}{2}\log|\textbf{K}|-\frac{\ell}{2}\log(2\pi)
\end{equation}
where $\theta = [\theta_1,...,\theta_{d_x}]^T$. Once the GP has been conditioned on $\mathcal{D}^{\ell}$, it computes the posterior distribution of $f$, $\mathcal{GP}_f^{\ell}$, at a set of $n$ new points $\mathcal{X}$ as shown in \eqref{eq:posterior_dist}:
\begin{equation} \label{eq:posterior_dist}
	\mathcal{GP}_f^{\ell}(\mathcal{X})\sim\mathcal{N}\left(m^{\ell}_f(\mathcal{X}), \Sigma^{\ell}_f(\mathcal{X})\right)
\end{equation}
where
\begin{subequations}\label{eq:GP_moments}
    \begin{align}
	m_f^\ell(\mathcal{X}) & = \textbf{K}(\mathcal{X}, x_{\mathcal{K}})^T\textbf{K}(x_{\mathcal{K}}, x_{\mathcal{K}})^{-1}f_{\mathcal{K}}\\[5pt]
	\Sigma^{\ell}_f(\mathcal{X}) & = \textbf{K}(\mathcal{X}, \mathcal{X})-\textbf{K}(\mathcal{X}, x_{\mathcal{K}})^T\textbf{K}(x_{\mathcal{K}}, x_{\mathcal{K}})^{-1}\textbf{K}(x_{\mathcal{K}}, \mathcal{X})
    \end{align}
\end{subequations}
\textcolor{black}{Note that the above formulation assumes that the observations in $\mathcal{D}^{\ell}$ are noise-free (i.e., system is perfectly observable). In many cases, however, data may be corrupted by noise, which can be represented as $f(x)=z(x)+e$. Here $z(x)$ is the true observation and $e$ is the noise. If $e$ follows a distribution of the form $\mathcal{N}\sim(0, \sigma^2_e)$, \eqref{eq:GP_moments} can be modified to account for the noise in the data:
\begin{subequations}\label{eq:GP_moments_noisy}
    \begin{align}
	m_f^\ell(\mathcal{X}) & = \textbf{K}(\mathcal{X}, x_{\mathcal{K}})^T\left[\textbf{K}(x_{\mathcal{K}}, x_{\mathcal{K}})+\sigma_e\textbf{I}\right]^{-1}f_{\mathcal{K}}\\[5pt]
	\Sigma^{\ell}_f(\mathcal{X}) & = \textbf{K}(\mathcal{X}, \mathcal{X})-\textbf{K}(\mathcal{X}, x_{\mathcal{K}})^T\left[\textbf{K}(x_{\mathcal{K}}, x_{\mathcal{K}})+\sigma_e\textbf{I}\right]^{-1}\textbf{K}(x_{\mathcal{K}}, \mathcal{X})
    \end{align}
\end{subequations}}
The mean and variance estimates obtained from \eqref{eq:GP_moments} or \eqref{eq:GP_moments_noisy} determine the expected performance and information gain of sampling at a new point and are passed into an acquisition function to determine the value of sampling at a new point $x$. In this work, we use the lower confidence bound \nomenclature{\textbf{LCB}}{Lower Confidence Bound acquisition function} (LCB) acquisition function, which has the form 
\begin{equation} \label{eq:acquisition_function}
	\mathcal{AF}^{\ell}(x)=m_f^\ell(x)-\kappa\cdot\sigma_f^\ell(x)
\end{equation}
where $\kappa\in\mathbb{R}_{+}$ is a hyperparameter, commonly referred to as the exploration weight, that determines the importance placed on the model uncertainty; larger values of $\kappa$ will make the algorithm more explorative while smaller values result in more exploitative behavior. The next sample point, $x^{\ell+1}$, is determined by solving the AF optimization problem defined in \eqref{eq:AF_opt}:
\begin{subequations}\label{eq:AF_opt}
	\begin{gather}
		x^{\ell+1} = \mathop{\textrm{argmin}}_{x} \mathcal{AF}^{\ell}(x;\kappa)\\
		\textrm{s.t.}~~x\in X
	\end{gather}
\end{subequations}
After taking a sample at $x^{\ell+1}$, the dataset is updated and the GP is retrained. This process is repeated until a satisfactory solution is found or the data collection budget is exhausted. The pseudocode for S-BO is presented in Algorithm \ref{alg:S_BO} and Figure \ref{fig:S_BO} provides an illustrative summary of this workflow.

\RestyleAlgo{ruled}
\begin{algorithm}[hbt!]
\caption{Standard Bayesian Optimization \textcolor{black}{(S-BO)}}\label{alg:S_BO}
Given $\kappa$, $L$, and $\mathcal{D}^{\ell}$\;
Train $\mathcal{GP}_f^{\ell}$ using initial dataset $\mathcal{D}^{\ell}$ and obtain $\mathcal{AF}^\ell$\;
\For{$\ell=1, 2,..., L$}{
Compute $x^{\ell+1} \gets \mathop{\textrm{argmin}}_{x}{\mathcal{AF}^{\ell}(x;\kappa)}$ s.t. $x\in X$\;
Sample system at $x^{\ell+1}$ to obtain $f^{\ell+1}$\;
Update dataset $\mathcal{D}^{\ell+1}\gets \mathcal{D}^{\ell}\cup  \left\{x^{\ell+1}, f^{\ell+1}\right\}$\;
Train GP using $\mathcal{D}^{\ell+1}$ to obtain $\mathcal{GP}_f^{\ell+1}$ and $\mathcal{AF}^{\ell+1}$\;
}
\end{algorithm}

\section{Bayesian Optimization with Composite Functions}

The use of a composite function objective in a BO setting was introduced in \cite{Astudillo:2019}. In this context, problem \eqref{eq:goal} is recast as:
\begin{subequations} \label{eq:composite_goal}
    \begin{gather}
	\min_x~~f(x, y(x))\\
	\textrm{s.t.}~~x\in X
    \end{gather}
\end{subequations}
where $f$ is now a known composite function with $f: X\times Y\to\mathbb{R}$, and $y:X\to\mathbb{R}^{d_y}$ is a black-box vector-valued function with range $Y\subseteq\mathbb{R}^{d_y}$ that captures the unknown intermediate elements of the system. Note that $y$ can be set up so that any element $y_i$ is only dependent on a subset of the inputs in $x$ or to have a nested structure where $y_i$ is also a function of another element in $y$, $y_j$ \cite{Astudillo:2021funnets, Paulson:2022, Xu:2023}. This feature makes this approach especially adept at representing complex systems where inputs often enter at different sections (e.g., material and energy inputs) and several of the elements in $y$ are interdependent (e.g., inter-unit streams, yields, recycle loops). As $f$ is now a known function, the formulation in \eqref{eq:composite_goal} shifts the modeling task from estimating the performance function to estimating the intermediate functions. In this work, we model $y(x)$ by using an independent single-output GP for each of the black-box elements. While multi-output GP models that can consider the correlation between outputs exist \cite{Alvarez:2012, Liu:2018}, these generally exhibit a higher computational complexity than single-output GPs and have a greater number of hyperparameters. Additionally, we can use the nested structure of $y$ to capture the correlation between relevant subcomponents $y_i$.
\\

The shift from black-box to a known composite function results in a loss of the direct performance and uncertainty estimates for $f$ that are available in S-BO. Instead, these must be inferred from the statistical moments of $y$ obtained via the GP model. For the case where $f$ is a linear transformation of $y$ of the form $f=a^Ty+b$, this can be done by making use of the closure of normal random variables under linear operations:
\begin{subequations}\label{eq:lin_f_moments}
    \begin{align}
	m_f^\ell(x) &  = a^Tm_y^{\ell}(x)+b\\[5pt]
	\sigma^{\ell}_f(x) & = \sqrt{a^T\Sigma^{\ell}_y(x)a},
    \end{align}
\end{subequations}
where $m_y^{\ell}(x)\in\mathbb{R}^{d_y}$ and $\Sigma_y^{\ell}(x)\in\mathbb{R}^{d_y\times d_y}$ are the mean and variance of $y$. However, in the more general case where $f$ is a nonlinear transformation, this analytical property no longer holds, and closed-form analytical expressions for $m_f^{\ell}(x)$ and $\sigma_f^{\ell}(x)$ are not readily available. Composite function BO paradigms have typically addressed this issue by using some variation of Monte Carlo sampling that allows for these values to be estimated numerically \cite{Astudillo:2021, Balandat:2020, Paulson:2022}. An alternative approach was presented in \cite{Xu:2023} wherein the GP models of $y(x)$ are used to construct a set of upper and lower confidence bound functions that enable the optimization problem to be cast onto a augmented space, $X\times \hat{Y}^{\ell}$, where $\hat{Y}^{\ell}$ is the range of the intermediate functions estimated by the confidence bounds. The details of these existing methods are discussed in Sections \ref{sec:mcbo} and \ref{sec:opbo}.

\subsection{Monte Carlo BO}\label{sec:mcbo}

Given the GP models of the intermediate functions $\mathcal{GP}_y^{\ell}$, trained on a dataset $\mathcal{D}_y^{\ell}=\{x_\mathcal{K}, y_\mathcal{K}\}$, Monte Carlo sampling estimates the mean and variance of the performance function at some point $x$ of interest by drawing $S$ samples from the distribution of $y$ generated by $\mathcal{GP}_y^{\ell}(x)$. These samples are then propagated via $f(x, y(x))$ and generate a range of outcomes that allow for the numerical estimation of $m_f^{\ell}(x)$ and $\sigma_f^{\ell}(x)$:
\begin{subequations}\label{eq:MC_moments}
    \begin{align}
	\hat{m}_f^{\ell}(x) & = \frac{1}{S}\sum_{s=1}^S f(x, m_y^{\ell}(x)+A_y^{\ell}(x)z_s)\\[5pt]
	\hat{\sigma}^{\ell}_f(x) & = \frac{1}{S-1}\sqrt{\sum_{s=1}^S \left(f(x, m^{\ell}_y(x)+A_y^{\ell}(x)z_s)-\hat{m}^{\ell}_f(x)\right)^2}
    \end{align}
\end{subequations}
Here, $A_y^{\ell}(x)\in\mathbb{R}^{d_y\times d_y}$ is the Cholesky factor of the GP covariance $(A_y^{\ell}(A_y^{\ell})^T=\Sigma_y^{\ell})$ and $z_s\in\mathbb{R}^{d_y}$ is a random vector drawn from $\mathcal{N}(\textbf{0}, \textbf{I})$. These estimates are used to construct the lower confidence bound for composite functions \nomenclature{\textbf{LCB-CF}}{Lower Confidence Bound for Composite Functions acquisition function} (LCB-CF) AF, $\mathcal{AF}_{\textrm{LCB-CF}}^{\ell}$, as outlined in Algorithm \ref{alg:LCB_CF}. As in S-BO, the AF is then optimized to select a select a new sampling point. The resulting data is appended to $\mathcal{D}_y^{\ell}$ and the GP models are retrained. The framework for this paradigm (which we refer to as \nomenclature{\textbf{MC-BO}}{Monte-Carlo Bayesian Optimization algorithm} MC-BO) is summarized in Algorithm \ref{alg:MC_BO}. Note that this is quite similar to S-BO, with the main differences being the shift to modeling the intermediate functions and the use of the LCB-CF. 
\\

\RestyleAlgo{ruled}
\begin{algorithm}[hbt!]
\caption{Lower Confidence Bound for Composite Functions \textcolor{black}{(LCB-CF)}}\label{alg:LCB_CF}
Given $x$, $\mathcal{GP}_y^{\ell}$, $S$, and $\kappa$\;
$m_y^{\ell}(x), \Sigma_y^{\ell}(x) \gets \mathcal{GP}_y^{\ell}(x)$\;
Calculate the Cholesky decomposition of $\Sigma_y^{\ell}(x)$ to determine the Cholesky factor $A_y^{\ell}(x)$\;
\For{$s=1, 2,..., S$}{
Draw sample $z_{s}$ from $\mathcal{N}(\textbf{0}, \textbf{I})$\;
$m_{f,s}^{\ell}(x) \gets f(x, m_y^{\ell}(x)+A_y^{\ell}(x)z_s)$\;
}
$\hat{m}_f^{\ell}(x) \gets \frac{1}{S}\sum_{s=1}^S m_{f,s}^{\ell}(x)$\;
$\hat{\sigma}^{\ell}_f(x) \gets \frac{1}{S-1}\sqrt{\sum_{s=1}^S \left(m_{f,s}^{\ell}(x)-\hat{m}^{\ell}_f(x)\right)^2}$\;
\Return{$\hat{m}_f^{\ell}(x)-\kappa\cdot\hat{\sigma}^{\ell}_f(x)$}
\end{algorithm}

\RestyleAlgo{ruled}
\begin{algorithm}[hbt!]
\caption{Monte Carlo Bayesian Optimization \textcolor{black}{(MC-BO)}}\label{alg:MC_BO}
Given $\kappa$, $L$, and $\mathcal{D}_y^{\ell}$\;
Train $\mathcal{GP}_y^\ell$ using initial dataset $\mathcal{D}_y^{\ell}$ and obtain $\mathcal{AF}_{\textrm{LCB-CF}}^{\ell}$\;
\For{$\ell=1, 2,..., L$}{
Compute $x^{\ell+1} \gets \mathop{\textrm{argmin}}_{x}{\mathcal{AF}_{\textrm{LCB-CF}}^{\ell}(x;\kappa)}$ s.t. $x\in X$\;
Sample system at $x^{\ell+1}$ to obtain $y^{\ell+1}$\;
Update dataset $\mathcal{D}_y^{\ell+1}\gets \mathcal{D}_y^{\ell}\cup  \left\{x^{\ell+1}, y^{\ell+1}\right\}$\;
Train GP using $\mathcal{D}_y^{\ell+1}$ to obtain $\mathcal{GP}_y^{\ell+1}$ and $\mathcal{AF}_{\textrm{LCB-CF}}^{\ell+1}$\;
}
\end{algorithm}

While Monte Carlo sampling provides a convenient approach for estimating the density of $f$, this is a computationally intensive method. Accurately estimating $m_f^{\ell}(x)$ and $\sigma_f^{\ell}(x)$ in regions of the design space with high model uncertainty or where $f(x, y(x))$ exhibits high sensitivity to variations in $y(x)$ can require a significant number of samples (on the order of $10^3$ or more). Given that the cost of drawing a sample from a GP scales as $\mathcal{O}(S\ell^3)$, generating the samples necessary in these instances can require a significant amount of computational time \cite{Shahriari:2016}. In addition, even though $f$ is a known function and is significantly cheaper to evaluate than the system, at large values of $S$ the computational cost of repeatedly evaluating $f(x, y(x))$ can also become nontrivial. This issue is compounded by the fact that \eqref{eq:MC_moments} must be recalculated at every point of interest. 

\subsection{Optimism-Driven BO}\label{sec:opbo}

\begin{figure*}[!h]
	\centering
	\includegraphics[width=0.75\textwidth]{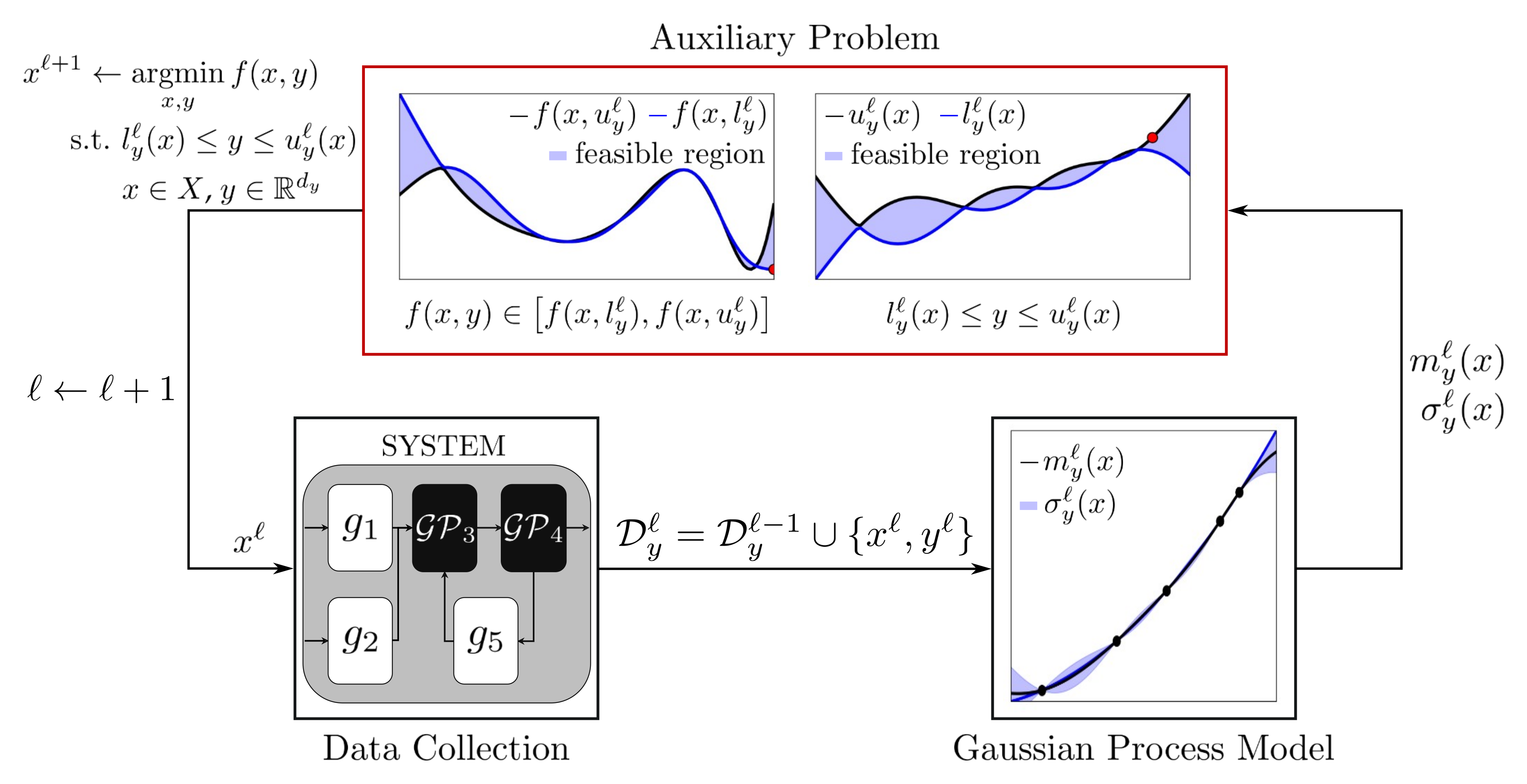}
	\caption{Workflow of the OP-BO algorithm. The data in $\mathcal{D}_y^{\ell}$ is used to construct $\mathcal{GP}_y^{\ell}$. The mean and uncertainty estimates calculated by the surrogate model are used to create a confidence interval bounded by $l_y^{\ell}(x)$ and $u_y^{\ell}(x)$ that constrains the possible values of $y$. These are incorporated into an auxiliary problem that is optimized to select a new sample point $x^{\ell+1}$. The resulting data is then appended to the dataset and the GP models are retrained}
	\label{fig:OP_BO}
\end{figure*}

When $f$ is formulated as a composite function, its derivatives can be calculated, making it possible to determine optimal values of $x$ and $y$ using gradient-based methods \cite{Urenholt:2019}. However, the solution might be infeasible as the proposed values of $y$ might be inconsistent with the relationships imposed by the intermediate functions. Typically, this issue is handled by using constraints that are designed to ensure feasibility. This requires that the closed-form representation of $y$ be available, which is not the case in composite function BO settings. However, the behavior of the intermediate functions can be estimated using $\mathcal{GP}_y^{\ell}$. The simplest approach for setting up the required constraints would then be to use the means of the GP models, but this discounts the information provided by the uncertainty estimates. The optimism-driven composite function BO algorithm (which we refer to as \nomenclature{\textbf{OP-BO}}{Optimism-driven Bayesian Optimization algorithm} OP-BO) proposes an alternative approach wherein the values of $y$ are instead restricted to a \textit{confidence interval} that is specified by the GP models \cite{Xu:2023}. This is done via a set of upper and lower confidence bound functions that are incorporated into the problem as inequality constraints and are of the form:
\begin{subequations}\label{eq:opbo_bounds}
    \begin{align}
	l_y^{\ell}(x) & = \max\{m_y^{\ell}(x)-\kappa\cdot\sigma_y^{\ell}(x), \hat{l}_y\}\\[5pt]
	u_y^{\ell}(x) & = \min\{m_y^{\ell}(x)+\kappa\cdot\sigma_y^{\ell}(x), \hat{u}_y\}
    \end{align}
\end{subequations}
where $\hat{l}_y\in\mathbb{R}^{d_y}$ and $\hat{u}_y\in\mathbb{R}^{d_y}$ are the lower and upper feasibility bounds of $y$, respectively; $\kappa$ determines the size of the confidence interval and thereby sets the emphasis placed on exploration similar to \eqref{eq:acquisition_function}. This allows OP-BO to construct and solve an auxiliary problem of \eqref{eq:composite_goal} of the form:
\begin{subequations} \label{eq:auxiliary_problem}
    \begin{align}
	\min_{x, y}~~&f(x, y)\\
	\textrm{s.t.}~~&l_y^{\ell}(x)-y \leq 0\\
	& y-u_y^{\ell}(x) \leq 0\\
	& x\in X, y\in\mathbb{R}^{d_y}
    \end{align}
\end{subequations}
This problem is solved at every iteration to determine the next sampling point $x^{\ell+1}$, filling the role of the AF in OP-BO. After sampling at this point, $\{x^{\ell+1}, y^{\ell+1}\}$ is appended to the current dataset and the GP models are re-trained allowing for \eqref{eq:opbo_bounds} to be updated. The workflow for the OP-BO algorithm is detailed in Algorithm \ref{alg:OP_BO} and Figure \ref{fig:OP_BO} provides an illustrative summary.
\\

\RestyleAlgo{ruled}
\begin{algorithm}[hbt!]
\caption{Optimism-Driven Composite Function Bayesian Optimization \textcolor{black}{(OP-BO)}}\label{alg:OP_BO}
Given $\kappa$, $L$, and $\mathcal{D}_y^{\ell}$\;
Train $\mathcal{GP}_y^\ell$ using initial dataset $\mathcal{D}_y^{\ell}$ and obtain $l_y^{\ell}$ and $u_y^{\ell}$\;
\For{$\ell=1, 2,..., L$}{
Compute $x^{\ell+1}$ by solving:\\
\[
   \hspace{-10em}
    \begin{aligned}
	\min_{x, y}~~&f(x, y) &\\
	\textrm{s.t.}~~&l_y^{\ell}(x)-y \leq 0 &\\
	& y-u_y^{\ell}(x) \leq 0 &\\
	& x\in X, y\in\mathbb{R}^{d_y} &
    \end{aligned}
\]\\ 
\smallskip
Sample system at $x^{\ell+1}$ to obtain $y^{\ell+1}$\;
Update dataset $\mathcal{D}_y^{\ell+1}\gets \mathcal{D}_y^{\ell}\cup  \left\{x^{\ell+1}, y^{\ell+1}\right\}$\;
Train GP using $\mathcal{D}_y^{\ell+1}$ to obtain $\mathcal{GP}_y^{\ell+1}$, $l_y^{\ell+1}$, and $u_y^{\ell+1}$\;
}
\end{algorithm}

Unlike the AFs used in S-BO and MC-BO, \eqref{eq:auxiliary_problem} does not require the estimation of the probability density of $f$. As such, OP-BO does not need to rely on the use of sampling methods. While this approach might seem more efficient, it should be noted that the dimensions of search space in the auxiliary problem ($\mathbb{R}^{d_x}+\mathbb{R}^{d_y}$) are larger than in the original problem ($\mathbb{R}^{d_x}$). As a result, problems that have a significant number of intermediate functions (large $d_y$) can lead to situations where \eqref{eq:auxiliary_problem} potentially requires a significant amount of computational time to solve. 


\section{The Bayesian Optimization for Interconnected Systems Approach}

While MC-BO and OP-BO provide distinct methodologies for handling composite functions in a BO setting, both paradigms are motivated by the same fundamental challenge: the lack of closed-form expressions for $m_f^{\ell}(x)$ and $\sigma_f^{\ell}(x)$. In the BOIS approach, we use derivatives of $f$ and the closure of Gaussian random variables under linear transformations to obtain an AF. 
\\

\begin{figure*}[!htp]
	\centering
	\includegraphics[width=0.5\textwidth]{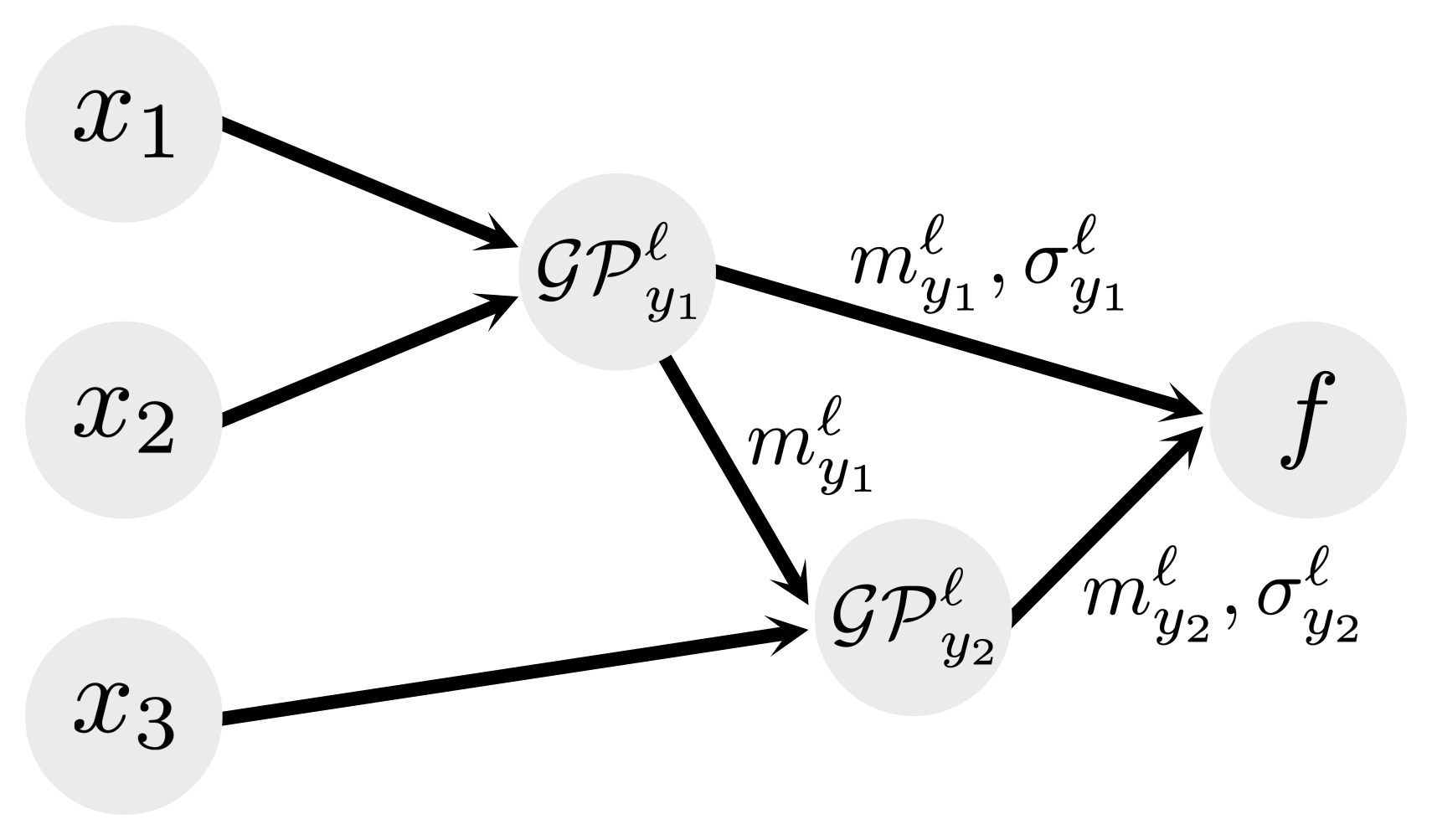}
	\caption{\textcolor{black}{Schematic representation of a nested function structure for $y$.}}
	\label{fig:nested_funs}
\end{figure*}

Consider the case where $f$ is a once-differentiable mapping with respect to $y$; in this scenario, it is possible to conduct a linearization of $f$ at the current iterate (as is done in standard optimization algorithms such as Newton's method). For the purpose of our discussion, we represent $f$ as:
\begin{equation} \label{eq:split_f}
	f(x, y(x)) = g(x)+h(x, y)
\end{equation}
Using a first-order Taylor series expansion, we linearize \eqref{eq:split_f} with respect to $y$ around a reference point $y_0$:
\begin{equation}\label{eq:linearization}
	f(x, y(x)) \approx g(x)+h(x, y_0)+J^T(y-y_0)
\end{equation}  
where the Jacobian is:
\begin{subequations}\label{eq:gradient}
    \begin{align}
	J & = \nabla_yh(x, y_0) \\
	  & = \nabla_yf(x, y_0).
    \end{align}
\end{subequations}
At a given point of interest, $x$, we calculate $m_y^{\ell}(x)$ and $\Sigma_y^{\ell}(x)$ using $\mathcal{GP}_y^{\ell}$. \textcolor{black}{Note that when $y$ exhibits a nested structure, any intermediates whose models depend on other elements in $y$ must be evaluated after their dependencies. This sequencing ensures that the current means of these inputs---values that subsequently are passed into these models---are available when the models are evaluated, as shown in Figure \ref{fig:nested_funs}. Once the full set of entries in $m_y^{\ell}(x)$ and $\Sigma_y^{\ell}(x)$ is obtained, we define the following:}
\begin{subequations}\label{eq:y_moments_bois}
    \begin{align}
	\Delta_\textrm{lo} &= \max\{0, \hat{l}_y-m_y^{\ell}(x)\} \label{eq:lower_clip} \\
	\Delta_\textrm{hi} &= \min\{0, \hat{u}_y-m_y^{\ell}(x)\} \label{eq:upper_clip} \\
	\hat{y}^{\ell} & = m_y^{\ell}(x)+\Delta_\textrm{lo}+\Delta_\textrm{hi} \\
	\hat{\Sigma}^{\ell} & = \Sigma_y^{\ell}(x)
    \end{align}
\end{subequations}
where \eqref{eq:lower_clip} and \eqref{eq:upper_clip} ensure that $\hat{y}^{\ell}$ is within any specified feasibility bounds of $y$. The rationale behind these calculations is that, if $m_y^{\ell}<\hat{l}_y$ or $m_y^{\ell}>\hat{u}_y$, then it is reasonable to interpret this as an indication that the true value of $y(x)$ is likely near the closest bound. If we then select a reference point, $\hat{y}_0^{\ell}$, in the $\epsilon$-neighborhood of $\hat{y}^{\ell}$, where $|\hat{y}_0^{\ell}-\hat{y}^{\ell}|\leq\epsilon$, we can approximate $f$ at $x$ as:
\begin{equation} \label{eq:f_example}
	f(x, y(x)) \approx g(x)+h(x, \hat{y_0}^{\ell})+J^T(y(x)-\hat{y}_0^{\ell})
\end{equation}
Note that, in this context, $g$ contains only the white-box elements of $f$ that have no dependency on $y$ and, therefore, $g(x)$ is a \textit{deterministic} variable. Combining this approximation of the performance function with \eqref{eq:lin_f_moments}, we are now able to derive a set of closed-form expressions that estimate the mean and uncertainty of $f$:
\begin{subequations}\label{eq:lin_moments}
    \begin{align}
	m_f^{\ell}(x) & = J^T\hat{y}^{\ell}+g(x)+h(x,\hat{y}_0^{\ell})-J^T\hat{y}_0^{\ell}\\[5pt]
	\sigma^{\ell}_f(x)& = \left(J^T\hat{\Sigma}^{\ell}J\right)^\frac{1}{2}
    \end{align}
\end{subequations}
\\

\begin{figure*}[!h]
	\centering
	\includegraphics[width=0.85\textwidth]{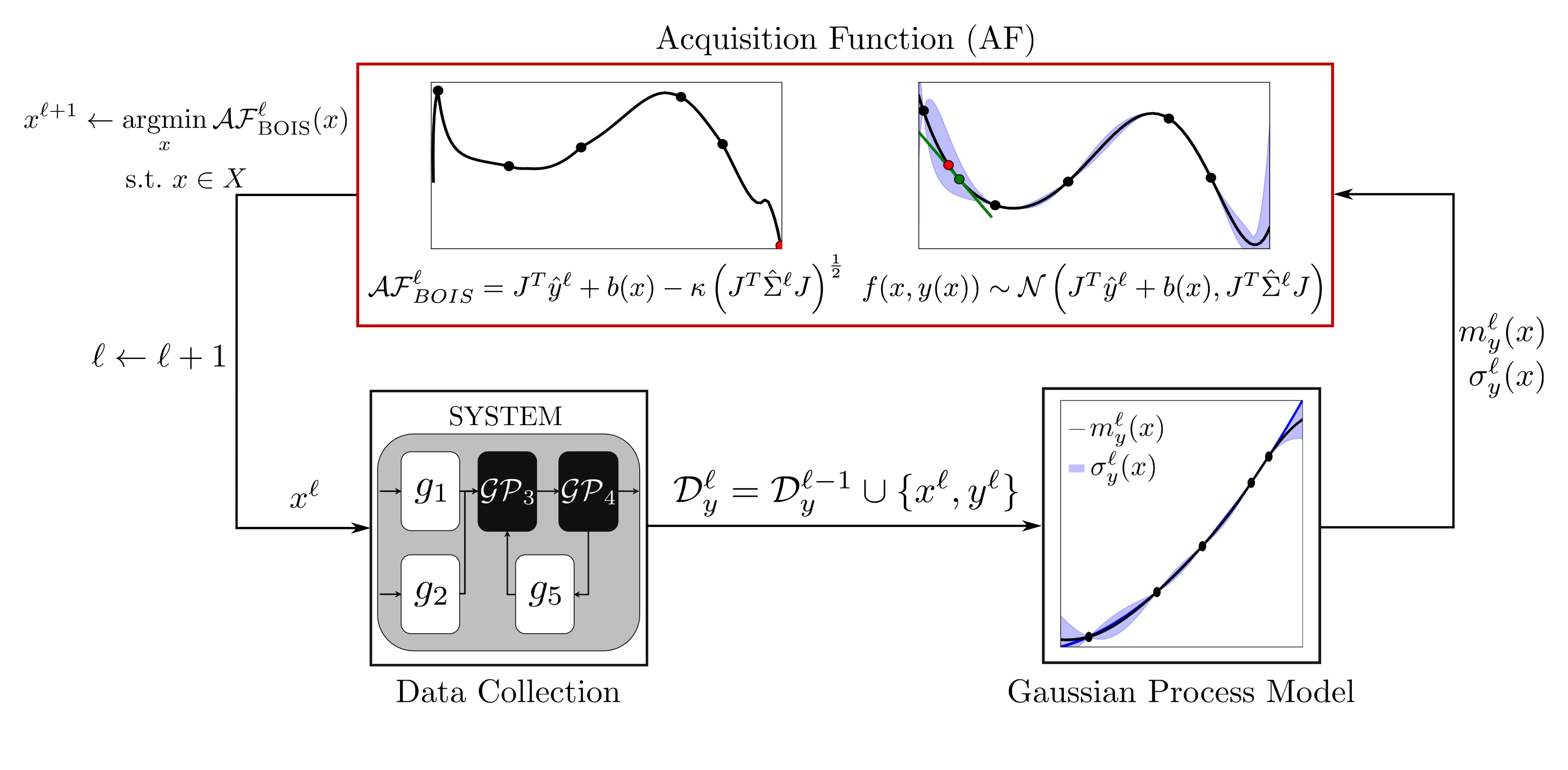}
	\caption{Workflow of the BOIS algorithm. Here, we note that $b(x)=g(x)+h(x,\hat{y}_0^{\ell})-J^T\hat{y}_0^{\ell}$. A set of GP surrogate models of $y$ is trained using $\mathcal{D}_y^{\ell}$. The mean and variance estimates calculated by the GPs are passed into $\mathcal{AF}_{\textrm{BOIS}}^{\ell}$ which generates a local Laplace approximation for the density of $f$. This AF is then optimized to obtain a new sample point, $x^{\ell+1}$. The system is then sampled at this point and the collected data is appended to the dataset and used to retrain the GP models.}
	\label{fig:BOIS}
\end{figure*}

The BOIS framework is initialized by using a dataset $\mathcal{D}_{y}^{\ell}$ to train a set of GP models of the intermediate functions $\mathcal{GP}_{y}^{\ell}$. Given a point of interest $x$ and the corresponding mean and uncertainty estimates for $y(x)$, \eqref{eq:linearization}-\eqref{eq:lin_moments} are used to construct the lower confidence bound for BOIS acquisition function \nomenclature{\textbf{LCB-BOIS}}{Lower Confidence Bound for BOIS acquisition function} (LCB-BOIS, denoted as $\mathcal{AF}_{\textrm{BOIS}}^{\ell}$), and this AF is then optimized to select a new sampling point, $x^{\ell+1}$. After sampling, the obtained datapoint $\{x^{\ell+1}, y^{\ell+1}\}$ is appended to the dataset and the GP models are retrained. A summary of this procedure is presented in Algorithm \ref{alg:BOIS} and Figure \ref{fig:BOIS} provides a visual representation. This framework is similar to MC-BO with the LCB-CF AF being replaced with LCB-BOIS.  

\RestyleAlgo{ruled}
\begin{algorithm}[hbt!]
\caption{Bayesian Optimization of Interconnected Systems \textcolor{black}{(BOIS)}}\label{alg:BOIS}
Given $\kappa$, $\epsilon$, $L$, and $\mathcal{D}_y^{\ell}$\;
Train $\mathcal{GP}_y^\ell$ using initial dataset $\mathcal{D}_y^{\ell}$ and obtain $\mathcal{AF}_{\textrm{BOIS}}^{\ell}$\;
\For{$\ell=1, 2,..., L$}{
Compute $x^{\ell+1} \gets \mathop{\textrm{argmin}}_{x}{\mathcal{AF}_{\textrm{BOIS}}^{\ell}(x;\kappa)}$ s.t. $x\in X$\;
Sample system at $x^{\ell+1}$ to obtain $y^{\ell+1}$\;
Update dataset $\mathcal{D}_y^{\ell+1}\gets \mathcal{D}_y^{\ell}\cup  \left\{x^{\ell+1}, y^{\ell+1}\right\}$\;
Train GP using $\mathcal{D}_y^{\ell+1}$ to obtain $\mathcal{GP}_y^{\ell+1}$ and $\mathcal{AF}_{\textrm{BOIS}}^{\ell+1}$\;
}
\end{algorithm}

Unlike MC-BO and OP-BO which are agnostic to the nature of the density of $f$, the BOIS framework implicitly assumes that $f$ is Gaussian in the neighborhood of the iterate $\hat{y}_0^{\ell}$. In other words, at any $x$ of interest, BOIS passes the mean and uncertainty estimates calculated by $\mathcal{GP}_y^{\ell}$ into \eqref{eq:f_example} to construct a local Laplace approximation of the performance function. As this approximation is then also Gaussian, it is also possible to obtain expressions for the probabilities and quantiles (to construct different types of AFs). Because $f(x, y(x))$ is likely not a normally distributed random variable, the Laplace approximation will result in a worse fit as the distance between $\hat{y}^{\ell}$ and $\hat{y}_0^{\ell}$ grows, similar to how \eqref{eq:linearization} becomes less accurate. \textcolor{black}{Thus, it is desirable to select a small value for $\epsilon$ to maximize the accuracy of the approximation. However, in cases where noise is present in the data, $\epsilon$ must be large enough to ensure that $\hat{y}^{\ell}$ and $\hat{y}_0^{\ell}$ are measurably different. It should also be mentioned that reductions in the value of $\epsilon$ beyond a certain point will only marginally improve the linear estimation and can potentially lead to numerical stability issues in the calculation of $J$. Note that BOIS does \textit{not} extrapolate the approximation generated at a previous point to estimate the distribution of $f$ at a new point. Instead, the linearization is updated in an adaptive manner. Specifically, at the current $x$, $\hat{y}^{\ell}$ and $\hat{\Sigma}^{\ell}$ are calculated from $\mathcal{GP}^{\ell}_y(x)$ and the corresponding $\hat{y}_0^{\ell}$ is obtained and $J$ is computed. These are then used to calculate $m^{\ell}_f(x)$ and $\sigma^{\ell}_f(x)$ as shown in \eqref{eq:lin_moments}.} 
\\

By deriving closed-from approximations for $m_f^{\ell}(x)$ and $\sigma_f^{\ell}(x)$, BOIS is able to reduce the number of function calls to $\mathcal{GP}_y^{\ell}$ and $f$ significantly when compared to MC-BO. At a given point $x$, BOIS only has to sample from the GPs once to obtain the estimates for $\hat{y}^{\ell}$ and $\hat{\Sigma}^{\ell}$, and the performance function is similarly only evaluated once to calculate $f(x, \hat{y}_0)$; recall that this is done tens to thousands of times in MC-BO. While BOIS does have to compute \eqref{eq:gradient}, this is also only done once at each iteration $x$. Additionally, computing function gradients has been shown to have a computational cost similar to that of evaluating the function itself when methods like automatic differentiation are used \cite{Griewank:2008}, \cite{Baur:1983}. As a result, the computational cost of obtaining $\mathcal{AF}_{\textrm{BOIS}}^{\ell}(x)$ can be significantly lower than that of calculating $\mathcal{AF}_{\textrm{LCB-CF}}^{\ell}(x)$. If we perform a similar comparison between BOIS and OP-BO, we observe that, like BOIS, OP-BO only samples from the GP models and evaluates the performance function once when setting up the auxiliary problem. However, the auxiliary problem is a constrained problem that is optimized over a higher dimensional space than the LCB-BOIS AF. As a result, OP-BO likely requires more computational time to obtain a new sample point than BOIS, especially when $y$ is high-dimensional.  


\section{Benchmark Studies}

We tested and compared the performances of S-BO, MC-BO, OP-BO and BOIS using various benchmark problems. Our aim is to demonstrate that BOIS can perform as well as or better than existing methods, while being less computationally intensive.  The data and code needed to reproduce the results can be found at \url{https://github.com/zavalab/bayesianopt}.
\\

The first study focuses on the performance of a simulated chemical process and the second study examines the design of a photobioreactor \nomenclature{\textbf{b-PBR}}{bag-Photobioreactor} (b-PBR) in a nutrient recovery process. In both systems, closed-form models are available for some of the process units, making them excellent candidates for benchmarking the composite function BO algorithms. A detailed overview of the systems used in each case study, along with the corresponding process unit models, can be found in the Supplementary Information (SI). 
\\

The MC-BO implementation used $S=100$ samples to calculate $\hat{m}_f^{\ell}(x)$ and $\hat{\sigma}_f^{\ell}(x)$, and the value of $\epsilon$ in BOIS was set to $\hat{y}^{\ell}\times 10^{-3}$. \textcolor{black}{Performance is reported either in terms of the raw cost or the log-normalized regret, $R$, which we define in \eqref{eq:log_regret} as:
\begin{equation}\label{eq:log_regret}
	R^{\ell} = \log_{10}\left(\left|\frac{f^{\ell}-f^{\ast}}{f^{\ast}}\right|\right)
\end{equation}
where $f^{\ast}$ is the true minimum value.}  
\\

All algorithms were implemented in Python 3.11 and used the {\tt gaussian\_process} module from Scikit-learn \cite{scikit-learn} for GP modeling. \textcolor{black}{Optimization of the AF was done via the SLSQP method with the {\tt minimize} function from Scipy \cite{scipy} using a multi-start at 50 randomly generated initial points to increase the probability of convergence to the global AF optimum. The gradient of $f$ in \eqref{eq:gradient} was evaluated using {\tt approx\_fprime}, also from Scipy. Differentiation issues arising from the activation of \eqref{eq:lower_clip} and \eqref{eq:upper_clip} were handled using subgradients, wherein $y_0$ is approached from the direction opposite to the active constraint when calculating $J$. This avoids the occurrence of zero-valued derivatives while also ensuring that the estimated gradient values are reasonable.}  

\subsection{Optimization of a Chemical Process}\label{subsec:rxt_flsh_rcyl}

Consider the following chemical process: reagents $A$ and $B$ are compressed, heated, and then fed into a reactor where they form product $C$. The reactor effluent is sent to a separator, where $C$ is recovered as a liquid. Note that $B$ is essentially non-condensable, while small amounts of $A$ can be present in the liquid phase. The vapor stream exiting the separator is largely composed of unreacted reagents. A fraction of this stream is recycled and fed back to the reactor after being heated and compressed, while the remainder is purged. The demand for $C$ is capped at a specified value $\bar{F}$ and any excess product generated cannot be sold. Our goal is to determine the operating temperatures and pressures of the reactor and separator as well as the recycle fraction that will minimize the operating cost of the process, which we define as:
\begin{subequations}\label{eq:objective_rxtr_flsh}
    \begin{align}
	f_1(x, y(x)) & = \sum_{j\in\{A,B\}}w_{j0}F_{j}+F_S\sum_{i\in\{A,B,C\}}w_{i}\psi_i+w_3\left(\frac{\psi_C F_S-\bar{F}}{\bar{F}}\right)^2\\[5pt]
	f_2(x,  y(x)) & = \sum_{h=1}^5 w_h\dot{Q}_h, +w_e\sum_{k=1}^3\dot{W}_k \\[5pt]
	f(x, y(x)) & = f_1(x, y(x))+f_2(x, y(x))
    \end{align}
\end{subequations}  
Here, $F_j$ denotes the molar flowrate of $A$ or $B$ into the process; $\psi_i$ is the mol fraction of $A$, $B$, or $C$ in the product stream which exits the process at rate $F_S$. The heating and cooling requirements of the heaters, reactor, and separator are denoted as $\dot{Q}_h$ and $\dot{W}_k$ is the power load of the $k^{\textrm{th}}$ compressor. The costs of reagents and heat and power utilities are $w_j$, $w_h$ and $w_e$ respectively, while $w_{i}$ refers to the value of species $i$ in the product stream. The demand cap is enforced via a quadratic penalty term that incurs an additional cost, scaled by $w_3$, when the process operates at value of $F_S$ that is different than $\bar{F}$.
\\

The design space was defined as the box domain $X = [673, 250, 288, 140, 0.5]\times[973, 450, 338,$ $170, 0.9]$, with the optimal solution of $-1890$ USD/hr located at $x = (844, 346, 288, 170, 0.9)$. \textcolor{black}{In the interest of providing the algorithms with sufficient samples to adequately search the design space while keeping the total computational time manageable (each function evaluation takes $\sim$1.2 minutes on average), we opted to measure algorithm performance across 25 trials.} During each trial, all algorithms ran for 100 iterations and were initialized using the same two points drawn from a uniform distribution of $X$. The reactor and separator were treated as black-boxes, and the compressors and heaters were assumed to be white-box elements. \textcolor{black}{This distinction was primarily based on the fact that the heater and compressor models are significantly simpler and faster to evaluate than the reactor and separator models.} 
\\

We defined the intermediate functions as the purge to feed ratio of $B$, $\eta_B$, the product to purge ratio of $A$, $\eta_A$, the purge to product ratio of $C$, $\eta_C$, and the utility requirements of the reactor and separator, $\dot{Q}_4$ and $\dot{Q}_5$ respectively. By combining these with the white-box models for the compressors and heaters, we were able to fully specify the system using only five intermediates. For comparison, if we had chosen to model the elements in \eqref{eq:objective_rxtr_flsh} directly, we would have had 8 black-box functions and we would not have been able to use the white-box models for the recycle compressor and heater. Additionally, by nesting some of the selected functions within each other, we were able to reduce the number of inputs used by the GP models of most of the intermediates. The specific details on the construction of the GP models for these can be found in the SI. 
\\

\begin{figure*}[!htp]
	\centering
	\includegraphics[width=1.0\textwidth]{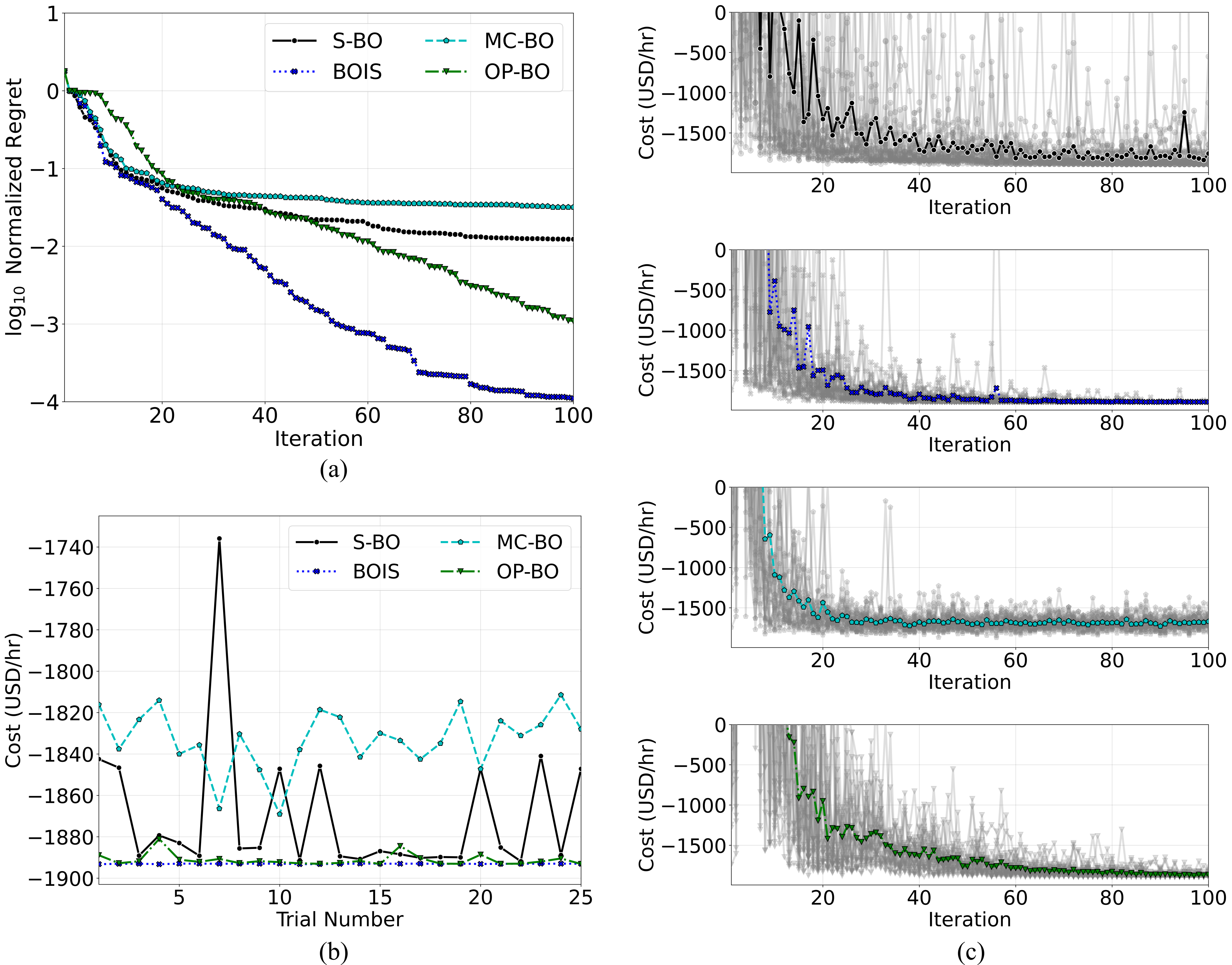}
	\caption{Performance comparison of the tested algorithms for the chemical process optimization problem based on (a) log-normalized regret of the best solution at the current iterate, (b) the best solution located by each algorithm during each trial, and (c) the distribution of the sampling behavior across the 25 runs for each of the tested methods with the average behavior shown in color.}
	\label{fig:results_1}
\end{figure*}

The results shown in Figure \ref{fig:results_1} summarize the performance of the tested algorithms across the 25 runs. We observed that BOIS outperformed the other methods. On average, it beat out  S-BO and MC-BO by 1.2\% and 3.3\% respectively in terms of solution value. While OP-BO returned a similarly valued solution, it required significantly more iterations to find it. BOIS was also remarkably robust, it consistently arrived at the global optimum regardless of where it was initialized. Again, OP-BO performed similarly, it located a solution within 0.1\% of the global optimum at every trial. S-BO and MC-BO exhibited significantly more variability, with S-BO appearing to be especially sensitive to the initial guess. Note that MC-BO was unable to find the global solution.
\\

The comparatively worse performances of S-BO and MC-BO were likely due to the fact that, as shown on the right side of Figure \ref{fig:results_1}, neither of these algorithms appeared to converge to a solution within 100 iterations. S-BO, especially, continued to sample from sub-optimal regions. This indicates that the algorithm struggled to learn the flow penalty and provides a clear demonstration of the advantages of employing a composite representation of $f$. The sampling behavior of the composite BO algorithms was significantly less variable as they were provided with a representation of the performance function that includes the flow penalty, allowing them to effectively identify the regions of the design space that minimize its value. S-BO, meanwhile, did not have access to this information and could only learn it by sampling, which was clearly an ineffective method.
\\

In the case of MC-BO, we surmise that its behavior is likely the result of the need for a greater number of samples. We observed that the performance function was sensitive to changes in the value of the intermediate functions. At a given point $x$ with $S=100$, different evaluations of the LCB-MCBO AF could return values that differed by over 10\%. This made the AF optimization step more likely to recommend a sub-optimal sampling point as it actually calculated a range of utility values rather than a single, replicable value as was the case in S-BO, OP-BO, and BOIS. While a solution to this problem would be to increase the value of $S$, this will also increase the computational cost of the algorithm. This highlights the advantage of utilizing the more specialized methods employed by OP-BO and BOIS to obtain a closed-form representation of the acquisition function/auxiliary problem over the more general Monte Carlo estimation approach.
\\
 
\begin{figure*}[!htp]
	\centering
	\includegraphics[width=1.0\textwidth]{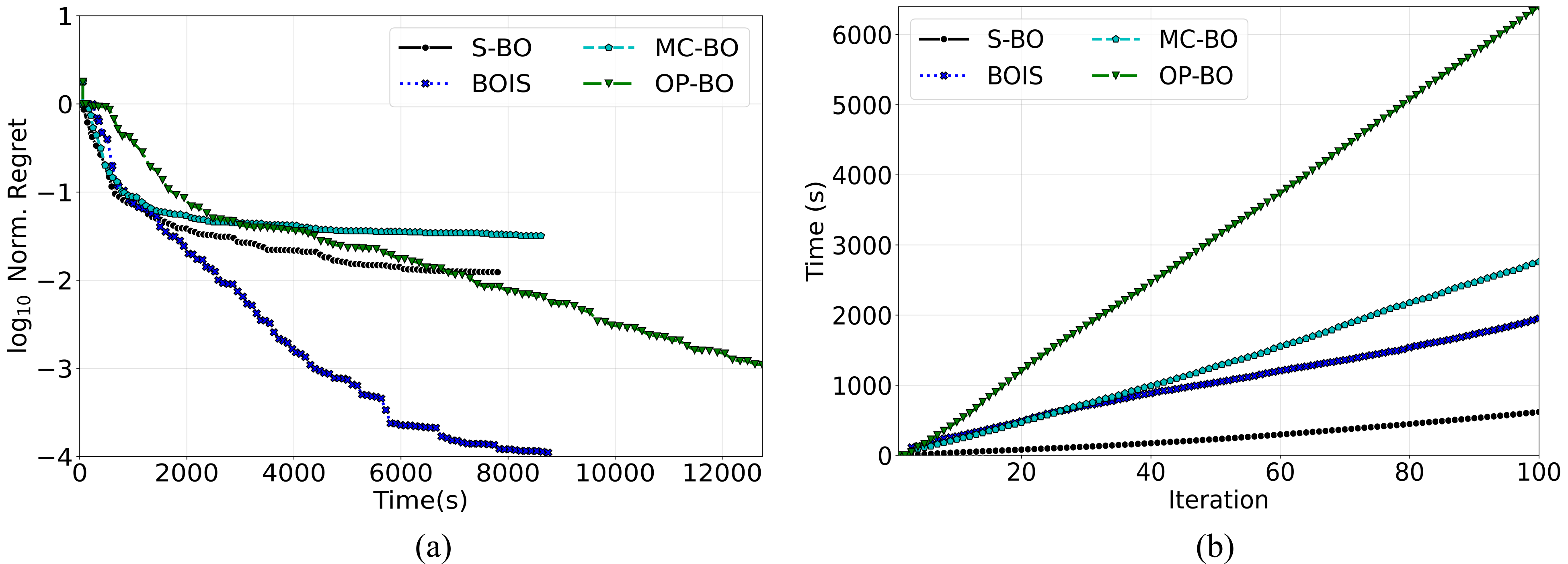}
	\caption{\textcolor{black}{Computational intensity of the tested algorithms for solving the chemical process optimization problem measured as (a) the total execution time and (b) the difference between the total execution time and total system sampling time.}}
	\label{fig:results_2}
\end{figure*}

\textcolor{black}{To evaluate the computational intensity of each algorithm, we measured their total execution time during each of the trials. The results shown on the left side of Figure \ref{fig:results_2} indicate that, on average, S-BO required the least amount of time to complete a trial (7800 seconds). This result is expected, as S-BO directly models the performance function and only needs to sample from its GP model to obtain the required mean and uncertainty estimates; this makes its AF evaluations faster. While these results might suggest that S-BO is the most efficient algorithm, it is important to also consider the quality of the solution. If we examine the amount of time each algorithm needed to be within 1\% of the global solution, we observe that BOIS and OP-BO reached this threshold after an average of 2,600 seconds and 7,400 seconds, respectively. Meanwhile, S-BO was unable to reach this value within the trial limits. This indicates that, while S-BO completed a trial faster, BOIS and OP-BO were able to find better solutions faster, as they are more effective at exploring the design space.}
\\ 

\textcolor{black}{We further quantified the variations in computational overhead by calculating the difference between the total execution time and total system sampling time for each algorithm. This metric is largely dominated by AF or auxiliary problem optimization step---the principal distinguishing feature between the tested methods. The results of this calculation, shown on the right side of Figure \ref{fig:results_2}, confirm that the AF optimization step in S-BO took noticeably less than in any of the composite function BO methods. Among the remaining algorithms, the AF optimization in BOIS was on average approximately 41\% faster than in MC-BO and 3.3 times faster than in OP-BO. This demonstrates that the methods we used to construct the LCB-BOIS AF were able to make it faster to evaluate and optimize than the LCB-CF and the OP-BO auxiliary problem. We are also able to observe how the efficiency of OP-BO can be significantly hampered due to the need to optimize the auxiliary problem over $x$ and $y$. As these are both five dimensional, solving \eqref{eq:auxiliary_problem} involves navigating a 10-dimensional space, significantly increasing the computational time required to solve this problem when compared to the AF optimization step of any of the other algorithms. Overall, these results underscore the importance of balancing solution value and computation time when selecting an optimizer. In cases where sampling from the system is the dominant bottleneck, the increased computational overhead of more advanced methods like BOIS can be justified. They are often able to find a satisfactory solution using significantly less samples, thereby reducing the overall time required. However, when this is not the case, the improvements in solution value may not be enough to justify the increase in resource use, and a simpler solution like S-BO might be a more efficient option}
\\

\begin{figure*}[!h]
	\centering
	\includegraphics[width=0.9\textwidth]{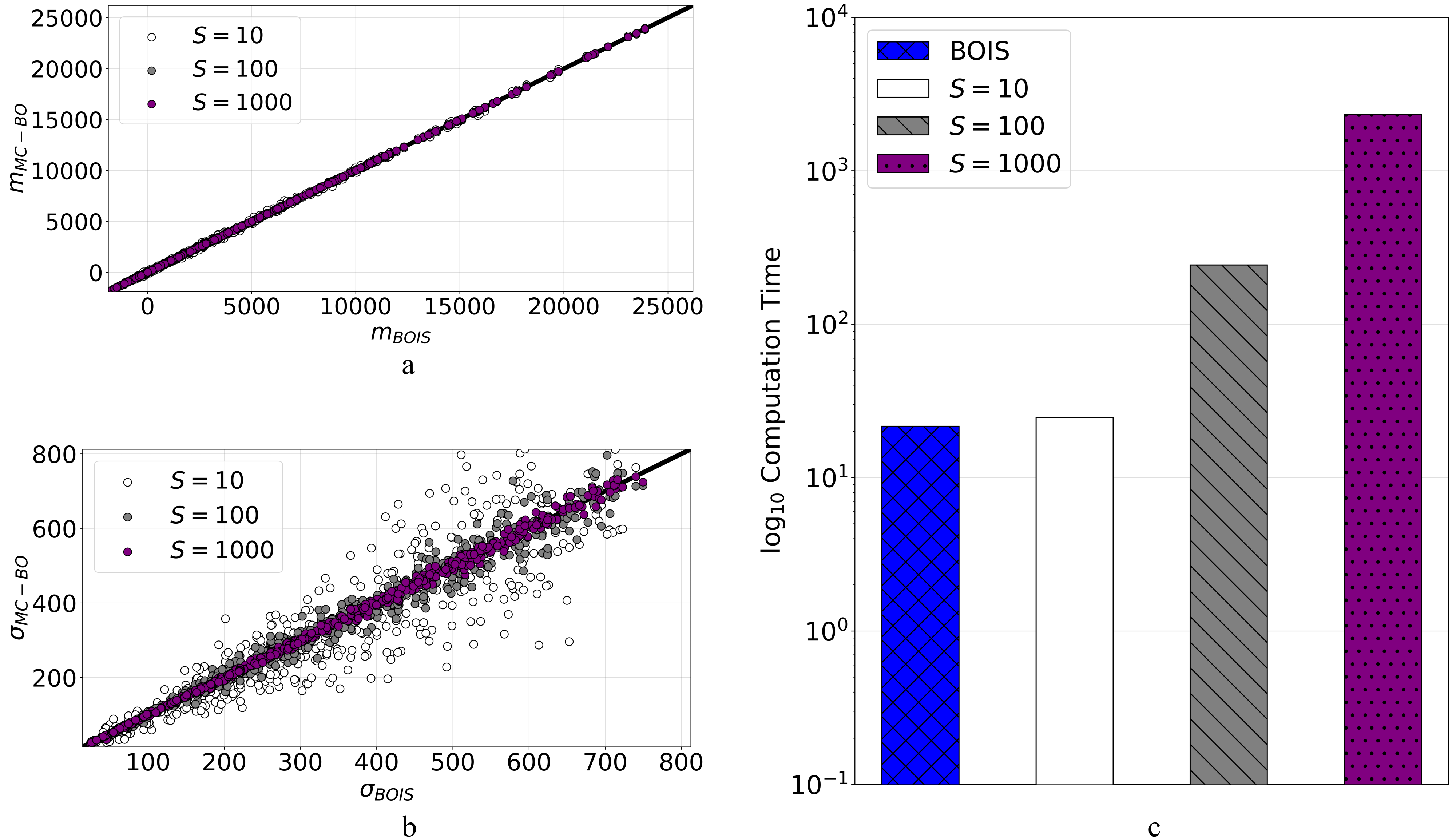}
	\caption{Parity plots of the estimates of $m_f^{\ell}(x)$  (a) and $\sigma_f^{\ell}(x)$ (b) for \eqref{eq:objective_rxtr_flsh} and $\log_{10}$ of the time required to generate the estimates (c) at 500 points in $X$ using BOIS with $\epsilon=\hat{y}\times 10^{-3}$ and MC-BO with samples sizes $S = 10$, $10^2$, and $10^3$; the same trained GP model of $y(x)$ was used by both algorithms.}
	\label{fig:results_3}
\end{figure*}

To confirm that BOIS provides accurate estimates of the mean and uncertainty of $f$, we compared the values calculated by BOIS for $m_f^{\ell}(x)$ and $\sigma_f^{\ell}(x)$ with those obtained from MC-BO. We know that as we increase the number of samples, \eqref{eq:MC_moments} will return values closer to the true moments of $f$. Using a trained GP model of $y(x)$, we calculated $\hat{m}_f^{\ell}(x)$ and $\hat{\sigma}_f^{\ell}(x)$ at 500 randomly selected points in $X$ using 10, 100, and 1,000 samples. If the values calculated by BOIS in \eqref{eq:lin_moments} are accurate, the difference between these values and those returned by MC-BO should decrease as $S$ increases. The results presented on the left side of Figure \ref{fig:results_3} demonstrate that this was precisely the case. While the estimates for $m_f^{\ell}(x)$ remained fairly constant across the various values of $S$, the estimates for $\sigma_f^{\ell}(x)$ were significantly more dynamic. We observed that BOIS estimated the uncertainty of $f$ with the same degree of accuracy as MC-BO with 1,000 samples. We also observed that the amount of time required to generate these estimates, shown on the right side of Figure \ref{fig:results_3}, was a couple of orders of magnitude lower when we used BOIS than when we used MC-BO with $S=1,000$. These results demonstrate that the adaptive linearization scheme employed by BOIS is not only fast but also accurate, further emphasizing that this method provides BOIS with a significant advantage over algorithms that rely on sampling-based techniques. 

\subsection{Optimization of a Photobioreactor}\label{subsec:pbr}

Nutrient management is a key challenge facing the agricultural sector as current practices are unsustainable. Processes that allow for nutrient recycling offer a potential solution to this issue. One such process involves the production of a cyanobacteria \nomenclature{\textbf{CB}}{Cyanobacteria} (CB) biofertilizer from animal waste. At the center of this operation is a bag photobioreactor (b-PBR) in which CB is grown. Due to the novelty of this application, this unit has not been widely studied and must be designed experimentally. However, computational methods like BO can aid in the identification of reactor settings that optimize overall process performance.
\\

In this case study, we considered the deployment of a biofertilizer production facility coupled with biogas generation using the waste produced at a hypothetical 1000 animal unit dairy farm. We measured performance using the minimum selling price \nomenclature{\textbf{MSP}}{Minimum Selling Price} (MSP) that the biofertilizer must be sold at to achieve a 15\% discounted return on investment \nomenclature{\textbf{DROI}}{Discounted Return On Investment} (DROI) over a 10-year project lifetime. \textcolor{black}{Due to the novelty of the CB cultivation system, detailed models for this unit operation have not been developed yet. The remaining units (biogas production and CB harvesting) are established technologies with models readily available in the literature (see \ref{supp-ap:pbr_model}). As a result, only the b-PBR was treated as a black-box.} Our goal was to identify the settings for three key reactor parameters---surface area to volume ratio ($\textrm{m}^{-1}$), batch time (days), and CB nutrient density (mass fraction)---that minimize the MSP. 
\\

We defined the design space as the box domain $X = [11.5, 22.5, 0.013]\times[19.2, 37.5, 0.154]$. Note that the MSP function is highly multi-modal within this domain and the global solution of $6.06$ USD/kg is located at $[15.4, 30, 0.0551]$. Given that the b-PBR was the only black-box in the system, it did not make sense to include all of the elements of the MSP function (material flows and unit sizes) in $y(x)$, as this would unnecessarily increase its dimensionality. Thus, we instead opted to use two intermediate functions that enable the full specification of the reactor: the total reactor volume, and the final CB titer (see \ref{supp-ap:pbr_intermediates}). In addition to reducing the size of $y(x)$, this selection allowed for the development of a b-PBR surrogate model that is highly refined in the regions near the optimum. The performance of each algorithm was measured across 125 trials, each initialized with a different pair of points selected from a $5\times 5\times 5$ grid of the design space. During each trial, all of the algorithms ran for 50 iterations and used the same set of initialization points. \textcolor{black}{As in the previous case study, this selection was primarily based on computational resource use considerations}
\\

\begin{figure*}[!htp]
	\centering
	\includegraphics[width=1.0\textwidth]{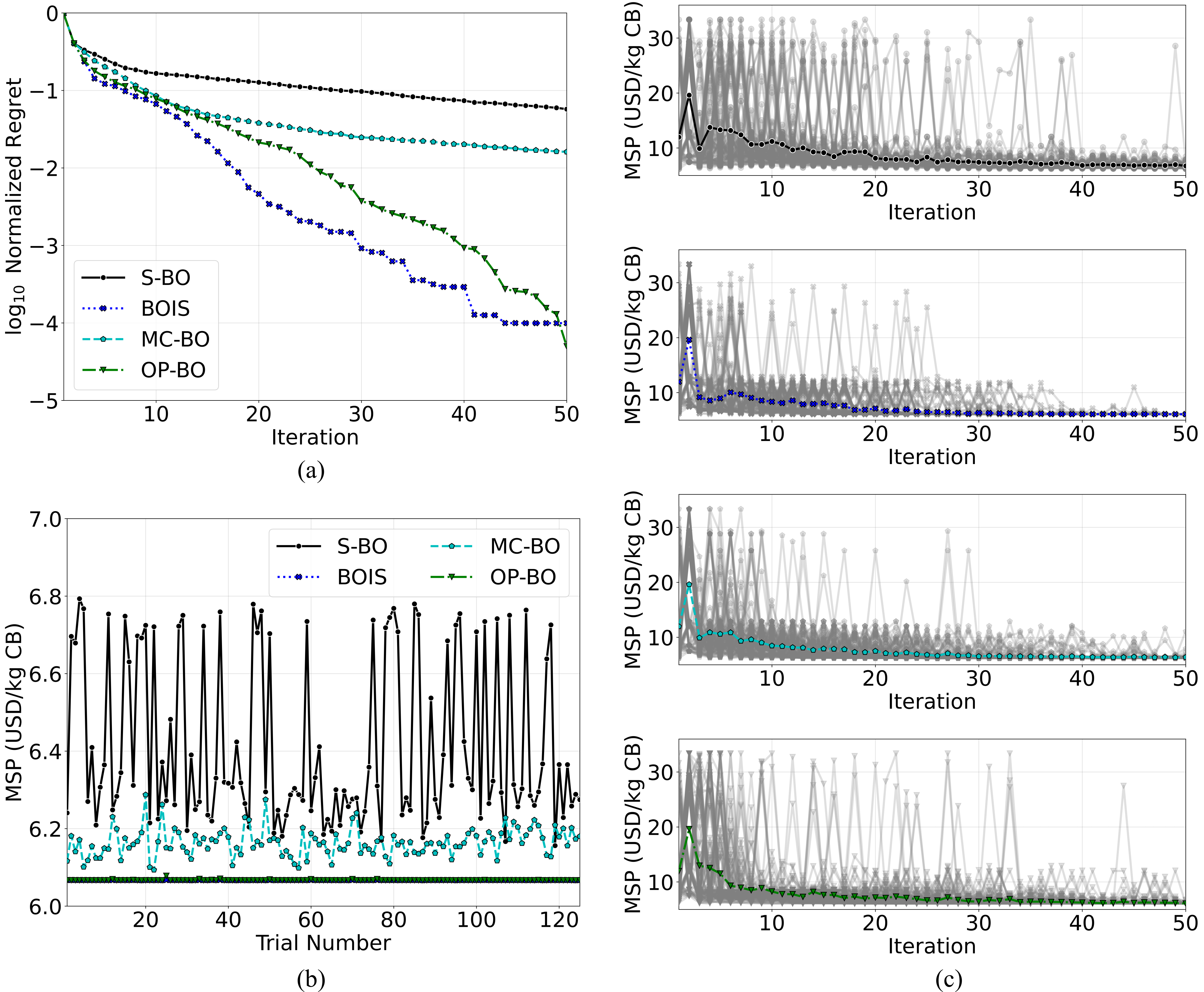}
	\caption{Performance comparison of the tested algorithms for the photobioreactor design problem based on (a) log-normalized regret of the best solution at the current iterate, (b) the best solution located by each algorithm during each trial, and (c) the distribution of the sampling behavior across the 125 runs for each of the tested methods with the average behavior shown in color.}
	\label{fig:results_4}
\end{figure*}

The performance profiles shown in Figure \ref{fig:results_4} illustrate that BOIS outperformed S-BO and MC-BO by and average 5.4\% and 1.6\% respectively. While OP-BO was able to locate the same optimum as BOIS, we observed that, on average, it took 10 additional iterations to find this point. BOIS was also the only algorithm that appeared to consistently converge by the end of each trial. While BOIS explored extensively during the first half of each trial, after approximately 40 iterations it tended to switch to exploiting the region near the optimum. Meanwhile, S-BO, MC-BO, and OP-BO continued to sample from sub-optimal regions, even at the end of each trial. From this, we can conclude that BOIS was able to navigate the design space more efficiently and could differentiate between optimal and sub-optimal regions more quickly than the other methods. Note that this increase in speed did not come at the expense of a decreased solution value as BOIS did not get trapped in a local optimum during any of the trials.
\\

In terms of robustness, we observed that OP-BO and BOIS were able to arrive at essentially the same solution regardless of where they were initialized. MC-BO exhibited some sensitivity, with best solution values located in each trial varying between -1.1\% to 2.0\% from the average minimum value. As was seen in the previous case study, this stems from the fact that the value returned by the LCB-CF AF at a given $x$ varies between evaluations. This can make it difficult to ascertain the true utility value of sampling at $x$, increasing the chance of selecting a sub-optimal sample point. However, it is worth noting that the variability observed for MC-BO was significantly less than what was observed for S-BO, which returned values across an almost 10\% range (-4.0\% to 5.9\%) around the average minimum value. This extreme sensitivity was likely due to the fact that the MSP is a difficult function to learn as it is highly non-smooth. As a result, S-BO was unable to construct a good surrogate model of the performance function, causing it to struggle to navigate the design space. Meanwhile, the composite function BO algorithms were tasked with learning functions that are comparatively much simpler and were thus better able to predict the behavior of the performance function. This demonstrates that, in addition to the structural system knowledge it provides, the ability to {\em shift the learning task from a complex function to a set of simpler and easier-to-learn intermediate functions} is a key advantage of using composite function BO over S-BO; this feature is part of the reason why MC-BO, OP-BO, and BOIS all outperformed S-BO.
\\

\begin{figure*}[!htp]
	\centering
	\includegraphics[width=1.0\textwidth]{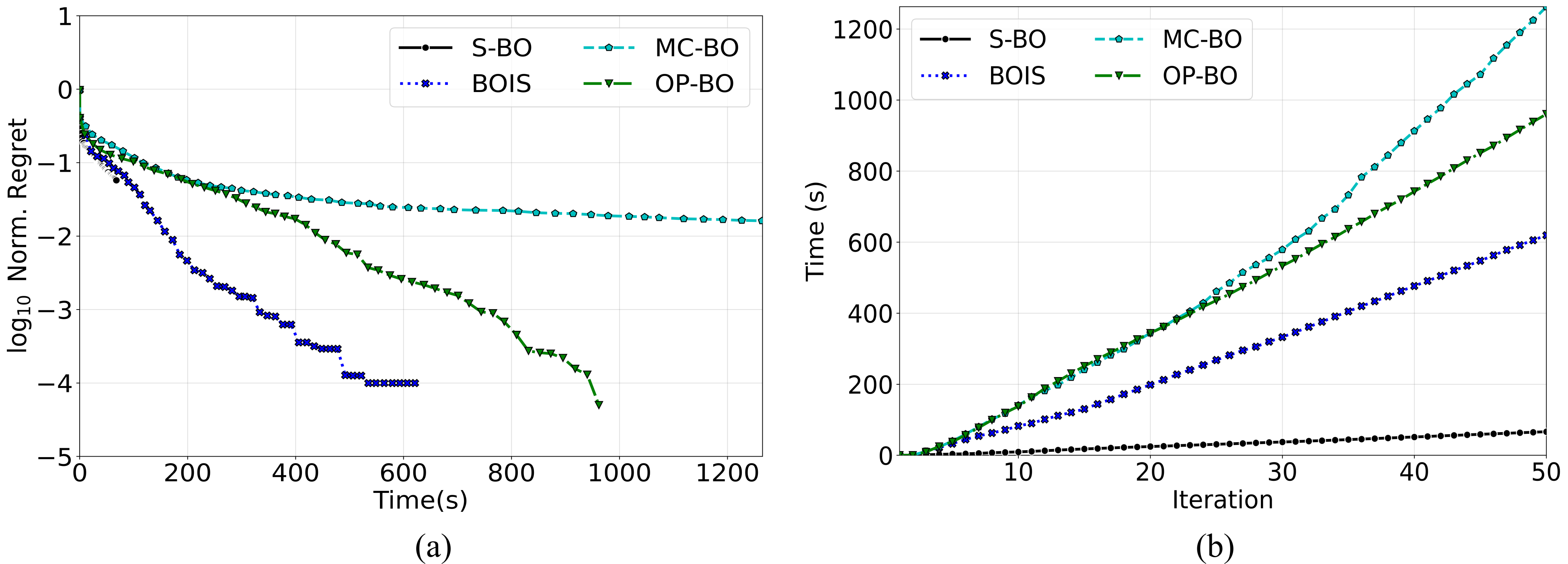}
	\caption{\textcolor{black}{Computational intensity of the tested algorithms for solving the photobioreactor design problem measured as (a) the total execution time and (b) the difference between the total execution time and total system sampling time.}}
	\label{fig:results_5}
\end{figure*}

\textcolor{black}{Figure \ref{fig:results_5} illustrates the computational intensity of the algorithms. For this case study, the reactors were modeled with a relatively simple preexisting model that was easy to evaluate. Due to the low system sampling cost, we observed that the computational intensity of the algorithms was largely dominated by the optimization of the acquisition function. As a result, S-BO was able to complete a trial approximately 10 times faster than its closest competitor on average. However, due to the comparatively poor value of the best solution S-BO found, the other algorithms were able to locate an improved solution using only roughly one-tenth the number of iterations. Consequently, despite having a significantly longer runtime per iteration, BOIS required only approximately 20 more seconds to locate a better solution. While these results highlight the aforementioned need to balance algorithm complexity with solution value, they also demonstrate that our proposed framework can be significantly more sample-efficient than S-BO. This might not be as relevant in the current context, but it becomes essential when considering the complexity required to build a reactor model from scratch. The experiments required for this task involve significant effort and can take days to complete \cite{Clark:2018}. Thus, if we aim to construct a more specialized model tailored to our application in the future, minimizing the number of iterations required to find an acceptable solution is critical for ensuring that BOIS is a practically useable tool.}
\\

\textcolor{black}{Moving to compare the performances of the composite function BO algorithms, we again observe that BOIS was the fastest method, outpacing OP-BO by 55\% and MC-BO by 104\%. Interestingly, unlike what was observed in \ref{subsec:rxt_flsh_rcyl}, OP-BO was faster than MC-BO. This underscores the fact that the comparative intensity of these methods is variable. At low values of $d_x$ and $d_y$, MC-BO is the more computationally expensive algorithm. However, as the dimensions of $x$ and $y$ increase, the time required to solve the auxiliary problem over the larger space increases to the point that the computational intensity of OP-BO becomes greater. Meanwhile, because BOIS utilizes a set of closed-form expressions to estimate $m_f^{\ell}(x)$ and $\sigma_f^{\ell}(x)$, it always requires fewer operations to evaluate its AF than MC-BO, and this AF is always optimized over a smaller space than the auxiliary problem in OP-BO. As a result, BOIS is able to maintain a consistent speed advantage over both MC-BO and OP-BO and appears to be a more scalable method.}
\\

\begin{figure*}[!h]
	\centering
	\includegraphics[width=0.9\textwidth]{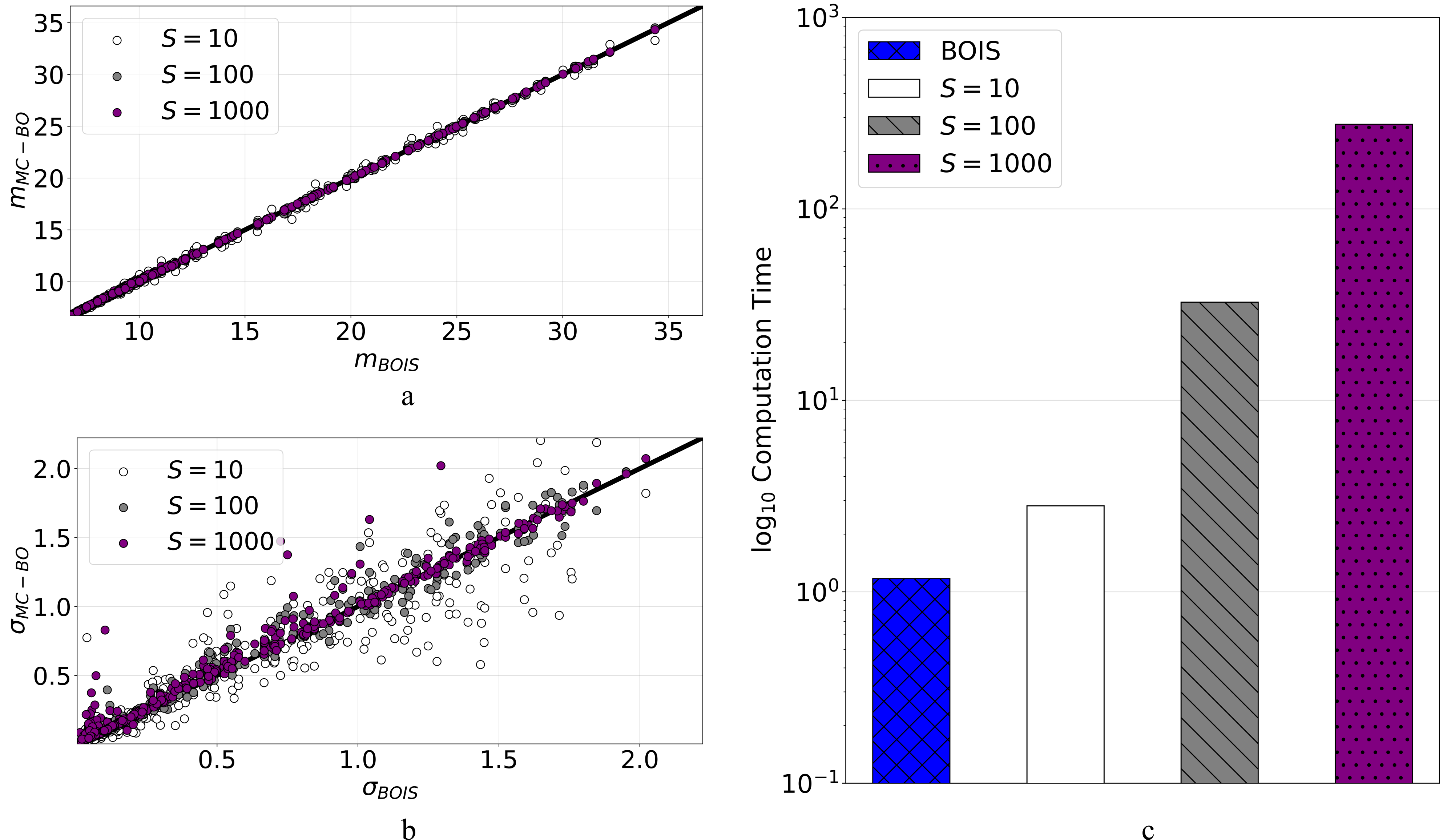}
	\caption{Parity plots of the estimates of $m_f^{\ell}(x)$ (a) and $\sigma_f^{\ell}(x)$ (b) for the MSP function and $\log_{10}$ of the time required to generate the estimates (c) at 500 points in $X$ using BOIS with $\epsilon=\hat{y}\times 10^{-3}$ and MC-BO with samples sizes $S = 10$, $10^2$, and $10^3$; the same trained GP model of $y(x)$ was used by both algorithms.}
	\label{fig:results_6}
\end{figure*}

Using the same approach as in chemical process optimization study, we estimated the values $m_f^{\ell}(x)$ and $\sigma_f^{\ell}(x)$ at 500 randomly selected points using BOIS and MC-BO. From these estimates, which are shown in Figure \ref{fig:results_6}, we can conclude that BOIS was once again able to generate highly accurate estimates of the statistical moments of the performance function. These results provide evidence that the adaptive linearization scheme is able to accurately estimate the behavior of a complex function like the MSP without necessarily requiring additional computational time. In fact, the relative differences in the generation times shown on the right side of Figure \ref{fig:results_6} were fairly similar to what was observed in the previous case study.
\\

If we specifically look at the spread of the estimates for $\sigma_f^{\ell}(x)$ calculated by BOIS versus those calculated by MC-BO when $S=1,000$, we observe that there appears to be a slight bias in the direction that the data points deviate from the center line. We believe that this is likely due to the fact that the intermediate functions are not actually symmetric as is assumed by the GP models (i.e., CB titer and reactor volume cannot be negative). This issue becomes especially poignant when $m_y^{\ell}(x)$ is near the feasibility bounds of any one of the intermediate functions, as the distribution of $y(x)$ spans values that are not permissible. We attempted to mitigate this problem by clipping the value of $m_y^{\ell}(x)$ to the corresponding upper or lower bound when it was outside of its allowable range as shown in \eqref{eq:y_moments_bois}. This provides BOIS with a workable solution as it is ensures that only feasible values of $\hat{y}^{\ell}$ and $\hat{y}_0^{\ell}$ are passed into $f(x, y(x))$. However, because MC-BO samples from the distribution to select the values of $y$, it can still select infeasible values. As a result, we had to clip sampled values of $y$ to $\hat{l}_y$ or $\hat{u}_y$ when they were outside of their permissible range. This caused the computed density of $f$ to not be symmetric and thus different from the density calculated by BOIS. While this indicates that the Laplace approximation does not provide an accurate representation of the density of $f$ near the bounds of the intermediate functions, we should note the numerical results we presented indicate that this was not an issue. This demonstrates that, despite assuming an incorrect shape, BOIS is still able to generate performance and informational value estimates that are at least consistent with those of the true underlying distribution. Additionally, it should be noted that at points where $m_y^{\ell}(x)$ was not near a boundary, which was the case for the majority of those selected, the approximation was still remarkably accurate.

\section{Conclusions and Future Work}

This work provides a detailed implementation of the BOIS framework for Bayesian Optimization (BO). BOIS is designed to facilitate the use of composite functions $f(x,y(x))$ in a BO setting. Composite performance functions offer an intuitive way for exploiting structural knowledge (in the form of physics or sparse system interconnections) and enable the integration of available white-box models. Additionally, this approach provides significant flexibility in selecting the black-box elements and setting up the corresponding surrogate models (i.e., the inputs can be customized). The key contribution of this work is the further development of this algorithm to handle feasibility considerations for the intermediate functions and the explicit consideration of white-box models via the customization of the intermediate functions guided by structural knowledge (e.g., mass and energy balances and process unit connectivity). This allows for a reduction in the dimensionality of $y(x)$ as well as in the inputs of the corresponding GP models, which improves the scalability of the algorithm. Additionally, we can specifically opt to consider intermediates of interest in order to develop surrogate models for these that are highly refined in regions around any explored optima. 
\\

We benchmarked the performance of BOIS against standard Bayesian optimization and existing composite function BO algorithms (MC-BO and OP-BO) using case studies arising in the context of chemical processes. Our results showed that BOIS significantly outperformed S-BO and was able to match or beat the performances of MC-BO and OP-BO while being significantly less computationally intensive. We also demonstrated that the values of the statistical moments of $f$ estimated by the adaptive linearization scheme we propose are generally very accurate and require significantly less time to compute compared to Monte Carlo estimates of comparable accuracy. However, we did observe a reduction in the accuracy of these predictions in regions near the feasibility limits of the intermediate functions. This is due to the symmetry assumption made by the GPs causing a significant portion of the calculated distribution of $y(x)$ to span non-permissible values in these regions. It should be noted, though, that BO is not limited to using GPs to construct the surrogate model; any probabilistic model can be used. Therefore, we would like to explore the use of alternatives such as warped GPs \cite{Snelson:2004}, RNNs \cite{Thompson:2023}, and reference models \cite{Lu:2021} as potential solutions to this issue. Additionally, we are also interested in investigating the performance of BOIS in higher dimensional systems and in developing alternative types of AFs that can extend the functionality of the algorithm, such as enabling parallelization. \textcolor{black}{Finally, we would like migrate our framework to a more comprehensive library, such as PyTorch or Jax. This will provide us with access to more advanced built-in features (e.g., auto-differentiation and heteroskedastic noise modeling) that can allows us to further improve the usability and efficiency of BOIS.}

\section{Supporting Information} 

Supplementary Information (SI) is available for this article. Included in the SI are detailed overviews of the systems used in each case study, along with the corresponding process unit models. The specific details on the construction of the GP models utilized can also be found in the SI. This information is available free of charge via the Internet at \url{https://pubs.acs.org}.

\section{Acknowledgments}

We acknowledge financial support from NSF-EFRI 2132036, the UW-Madison GERS program, and the PPG Fellowship. 

\printnomenclature

\bibliographystyle{abbrv}
\bibliography{./root}

\begin{thebibliography}{10}

\bibitem{Alvarez:2012}
M.~A. Alvarez, L.~Rosasco, and N.~D. Lawrence.
\newblock Kernels for vector-valued functions: a review.
\newblock {\em arXiv preprint arXiv:1106.6251}, 2012.

\bibitem{Astudillo:2019}
R.~Astudillo and P.~Frazier.
\newblock {B}ayesian optimization of composite functions.
\newblock In K.~Chaudhuri and R.~Salakhutdinov, editors, {\em Proceedings of
  the 36th International Conference on Machine Learning}, volume~97 of {\em
  Proceedings of Machine Learning Research}, pages 354--363. PMLR, 09--15 Jun
  2019.

\bibitem{Astudillo:2021funnets}
R.~Astudillo and P.~Frazier.
\newblock Bayesian optimization of function networks.
\newblock {\em Advances in neural information processing systems},
  34:14463--14475, 2021.

\bibitem{Astudillo:2021}
R.~Astudillo and P.~Frazier.
\newblock Thinking inside the box: {A} tutorial on grey-box {Bayesian}
  optimization.
\newblock In {\em Proceedings of the 2021 Winter Simulation Conference},
  December 2021.

\bibitem{Bajaj:2018}
I.~Bajaj, S.~S. Iyer, and M.~F. Hasan.
\newblock A trust region-based two phase algorithm for constrained black-box
  and grey-box optimization with infeasible initial point.
\newblock {\em Computers \& Chemical Engineering}, 116:306--321, 2018.

\bibitem{Balandat:2020}
M.~Balandat, B.~Karrer, D.~Jiang, S.~Daulton, B.~Letham, A.~Wislon, and
  E.~Bashky.
\newblock {BOTORCH}: {A} framework for efficient {Monte-Carlo} {Bayesian}
  optimization.
\newblock In {\em Proceedings of the 34th Conference International Conference
  on Neural Information Processing Systems}, NIPS '20, pages 21524--21538.
  Curran Associates Inc., Dec 2020.

\bibitem{Baur:1983}
W.~Baur and V.~Strassen.
\newblock The complexity of partial derivatives.
\newblock {\em Theoretical Computer Science}, 22(3):317--330, 1983.

\bibitem{Beykal:2018}
B.~Beykal, F.~Boukouvala, C.~A. Floudas, and E.~N. Pistikopoulos.
\newblock Optimal design of energy systems using constrained grey-box
  multi-objective optimization.
\newblock {\em Computers \& Chemical Engineering}, 116:488--502, 2018.

\bibitem{boukouvala:2017}
F.~Boukouvala and C.~Floudas.
\newblock {ARGONAUT}: {A}lgo{R}ithms for {G}lobal {O}ptimization of
  co{N}str{A}ined grey-box comp{UT}ational problems.
\newblock {\em Optimization Letters}, 11(5):895--913, 2017.

\bibitem{Brochu:2010}
E.~Brochu, V.~M. Cora, and N.~De~Freitas.
\newblock A tutorial on {Bayesian} optimization of expensive cost functions,
  with application to active user modeling and hierarchical reinforcement
  learning.
\newblock {\em arXiv preprint arXiv:1012.2599}, 2010.

\bibitem{Chakrabarty:2023}
A.~Chakrabarty, S.~A. Bortoff, and C.~R. Laughman.
\newblock Simulation failure-robust {Bayesian} optimization for data-driven
  parameter estimation.
\newblock {\em IEEE Transactions on Systems, Man, and Cybernetics: Systems},
  53(5):2629--2640, 2023.

\bibitem{Clark:2018}
R.~L. Clark, L.~L. McGinley, H.~M. Purdy, T.~C. Korosh, J.~L. Reed, T.~W. Root,
  and B.~F. Pfleger.
\newblock Light-optimized growth of cyanobacterial cultures: growth phases and
  productivity of biomass and secreted molecules in light-limited batch growth.
\newblock {\em Metabolic engineering}, 47:230--242, 2018.

\bibitem{Conn:2009}
A.~Conn, K.~Scheinberg, and L.~Vicente.
\newblock {\em Introduction to Derivative-free Optimization}.
\newblock SIAM, 2009.

\bibitem{Garnett:2023}
R.~Garnett.
\newblock {\em Baysian Optimization}.
\newblock Cambridge University Press, 2023.

\bibitem{Gonzalez:2023}
L.~D. González and V.~M. Zavala.
\newblock {BOIS: Bayesian Optimization of Interconnected Systems}.
\newblock {\em arXiv preprint arXiv:2311.11254}, 2023.

\bibitem{Greenhill:2020}
S.~Greenhill, S.~Rana, S.~Gupta, P.~Vellanki, and S.~Venkatesh.
\newblock Bayesian optimization for adaptive experimental design: {A} review.
\newblock {\em IEEE access}, 8:13937--13948, 2020.

\bibitem{Griewank:2008}
A.~Griewank and A.~Walther.
\newblock {\em Evaluating Derivatives: Principles and Techniques of Algorithmic
  Differentiation}.
\newblock SIAM, Philadelphia, 2008.

\bibitem{Hase:2021}
F.~Hase, M.~Aldeghi, R.~J. Hickman, L.~M. Roch, and A.~Aspuru-Guzik.
\newblock Gryffin: An algorithm for {Bayesian} optimization of categorical
  variables informed by expert knowledge.
\newblock {\em Applied Physics Reviews}, 8(3):031406, 2021.

\bibitem{Jones:1998}
D.~R. Jones, M.~Schonlau, and W.~J. Welch.
\newblock Efficient global optimization of expensive black-box functions.
\newblock {\em Journal of Global Optimization}, 13(4):455--492, 1998.

\bibitem{Kandasamy:2017}
K.~Kandasamy, G.~Dasarathy, J.~Schnieder, and B.~Pózcos.
\newblock Multi-fidelity {Bayesian} optimisation with continuous
  approximations.
\newblock In D.~Precup and Y.~Teh, editors, {\em Uncertainty in Artificial
  Intelligence}, volume~70 of {\em Proceedings of Machine Learning Research},
  pages 1799--1808. PMLR, 06--11 Aug 2017.

\bibitem{Lam:2018}
R.~R. Lam, M.~Poloczek, P.~I. Frazier, and K.~E. Willcox.
\newblock Advances in {Bayesian} optimization with applications in aerospace
  engineering.
\newblock In {\em Proceedings of the AIAA/ASCE/AHS/ASC Structures, Structural
  Dynamics, and Materials Conference}, Boston, MA, 2018. AIAA.

\bibitem{Liu:2018}
H.~Liu, J.~Cai, and Y.-S. Ong.
\newblock Remarks on multi-output {Gaussian} process regression.
\newblock {\em Knowedge-Based Systemsl}, 144:102--121, 2018.

\bibitem{Lu:2023}
C.~Lu and J.~A. Paulson.
\newblock No-regret constrained {Bayesian} optimization of noisy and expensive
  hybrid models using differentiable quantile function approximations.
\newblock {\em Journal of Process Control}, 131:103085, 2023.

\bibitem{Lu:2021}
Q.~Lu, L.~Gonz\'alez, R.~Kumar, and V.~Zavala.
\newblock Bayesian optimization with reference models: A case study in {MPC}
  for {HVAC} central plants.
\newblock {\em Computers \& Chemical Engineering}, 154:107491, 2021.

\bibitem{Matern:1960}
B.~Mat{\'e}rn.
\newblock Spatial variation : Stochastic models and their application to some
  problems in forest surveys and other sampling investigations.
\newblock In {\em Messages from the State Forestry Research Institute},
  volume~49, 1960.

\bibitem{Mockus:2012}
J.~Mockus.
\newblock {\em Bayesian Approach to Global Optimization: Theory and
  Applications}.
\newblock Springer Science \& Business Media, 2012.

\bibitem{Paulson:2022}
J.~Paulson and C.~Lu.
\newblock {COBALT}: {CO}nstrained {Bayesian} optimiz{A}tion of computaiona{L}ly
  expensive grey-box models exploiting deriva{T}ive information.
\newblock {\em Computers \& Chemical Engineering}, 160:107700, 2022.

\bibitem{scikit-learn}
F.~Pedregosa, G.~Varoquaux, A.~Gramfort, V.~Michel, B.~Thirion, O.~Grisel,
  M.~Blondel, P.~Prettenhofer, R.~Weiss, V.~Dubourg, J.~Vanderplas, A.~Passos,
  D.~Cournapeau, M.~Brucher, M.~Perrot, and E.~Duchesnay.
\newblock Scikit-learn: Machine learning in {P}ython.
\newblock {\em Journal of Machine Learning Research}, 12:2825--2830, 2011.

\bibitem{Priem:2019}
R.~Priem, N.~Bartoli, and Y.~Diouane.
\newblock On the use of upper trust bounds in constrained {Bayesian}
  optimization infill criterion.
\newblock In {\em AIAA Aviation 2019 Forum}, pages 1--10, Dallas, United
  States, June 2019.

\bibitem{Radivojevic:2020}
T.~Radivojević, Z.~Costello, K.~Workman, and H.~Garcia~Martin.
\newblock A machine learning automated recommendation tool for synthetic
  biology.
\newblock {\em Nature Communications}, 11(1):4879, 2020.

\bibitem{Rasmussen:2006}
C.~E. Rasmussen and C.~K.~I. Williams.
\newblock {\em Gaussian Processes for Machine Learning}.
\newblock MIT Press, 2006.

\bibitem{Shahriari:2016}
B.~Shahriari, K.~Swersky, Z.~Wang, R.~Adams, and N.~de~Freitas.
\newblock Taking the human out of the loop: {A} review of {Bayesian}
  optimization.
\newblock {\em Proceedings of the IEEE}, 104:148--175, 2016.

\bibitem{Snelson:2004}
E.~Snelson, Z.~Ghahramani, and C.~Rasmussen.
\newblock Warped {Gaussian} processes.
\newblock In S.~Thrun, L.~Saul, and B.~Sch\"{o}lkopf, editors, {\em Advances in
  Neural Information Processing Systems}, volume~16, pages 337--334. MIT Press,
  2004.

\bibitem{Snoek:2012}
J.~Snoek, H.~Larochelle, and R.~P. Adams.
\newblock Practical {Bayesian} optimization of machine learning algorithms.
\newblock In {\em Advances in Neural Information Processing Systems}, 2012.

\bibitem{Sohlberg:2008}
B.~Sohlberg and E.~Jacobsen.
\newblock Grey-box modeling – branches and experiences.
\newblock {\em IFAC Proceedings Volumes}, 41(2):11415--11420, 2008.
\newblock 17th IFAC World Congress.

\bibitem{Sorourifar:2023}
F.~Sorourifar, N.~Choksi, and J.~A. Paulson.
\newblock Computationally efficient integrated design and predictive control of
  flexible energy systems using multi-fidelity simulation-based {Bayesian}
  optimization.
\newblock {\em Journal of Optimal Control Applications and Methods},
  44(4):549--576, 2023.

\bibitem{Sorourifar:2021}
F.~Sorourifar, G.~Makrygirgos, A.~Mesbah, and J.~A. Paulson.
\newblock A data-driven automatic tuning method for {MPC} under uncertainty
  using constrained {Bayesian} optimization.
\newblock {\em IFAC-PapersOnLine}, 54(3):243--250, 2021.
\newblock 16th IFAC Symposium on Advanced Control of Chemical Processes ADCHEM
  2021.

\bibitem{Thompson:2023}
J.~Thompson, V.~Zavala, and O.~Venturelli.
\newblock Integrating a tailored recurrent neural network with {Bayesian}
  experimental design to optimize microbial community functions.
\newblock {\em PLOS Computational Biology}, 19(9):1--25, 2023.

\bibitem{Urenholt:2019}
A.~Uhrenholt and B.~Jensen.
\newblock Efficient {Bayesian} optimization for target vector estimation.
\newblock In K.~Chaudhuri and M.~Sugiyama, editors, {\em Proceedings of the
  Twenty-Second International Conference on Artificial Intelligence and
  Statistics}, volume~89 of {\em Proceedings of Machine Learning Research},
  pages 2661--2670. PMLR, 16--18 Apr 2019.

\bibitem{scipy}
P.~Virtanen, R.~Gommers, T.~E. Oliphant, M.~Haberland, T.~Reddy, D.~Cournapeau,
  E.~Burovski, P.~Peterson, W.~Weckesser, J.~Bright, S.~J. {van der Walt},
  M.~Brett, J.~Wilson, K.~J. Millman, N.~Mayorov, A.~R.~J. Nelson, E.~Jones,
  R.~Kern, E.~Larson, C.~J. Carey, {\.I}.~Polat, Y.~Feng, E.~W. Moore,
  J.~{VanderPlas}, D.~Laxalde, J.~Perktold, R.~Cimrman, I.~Henriksen, E.~A.
  Quintero, C.~R. Harris, A.~M. Archibald, A.~H. Ribeiro, F.~Pedregosa, P.~{van
  Mulbregt}, and {SciPy 1.0 Contributors}.
\newblock {{SciPy} 1.0: Fundamental Algorithms for Scientific Computing in
  Python}.
\newblock {\em Nature Methods}, 17:261--272, 2020.

\bibitem{Wilson:2014}
A.~Wilson, A.~Fern, and P.~Tadepalli.
\newblock Using trajectory data to improve {Bayesian} optimization for
  reinforcement learning.
\newblock {\em Journal of Machine Learning Research}, 15(8):253--282, 2014.

\bibitem{Wu:2021}
J.~Wu, S.~Toscano-Palmerin, P.~I. Frazier, and A.~G. Wilson.
\newblock Practical multi-fidelity {Bayesian} optimization for hyperparameter
  tuning.
\newblock {\em arXiv preprint arXiv:1903.04703}, 2019.

\bibitem{Xu:2023}
W.~Xu, Y.~Jiang, B.~Svetozarevic, and C.~N. Jones.
\newblock Bayesian optimization of expensive nested grey-box functions.
\newblock {\em arXiv preprint arXiv:2306.05150}, 2023.

\end{thebibliography}


\begin{thebibliography}{10}

\bibitem{villegas:2019}
H.~A. Aguirre-Villegas, R.~A. Larson, and M.~A. Sharara.
\newblock Anaerobic digestion, solid-liquid separation, and drying of dairy
  manure: Measuring constituents and modeling emission.
\newblock {\em Science of the Total Environment}, 696:134059, 2019.

\bibitem{cepci:2020}
{Chemical Engineering}.
\newblock {Chemical Engineering Plant Cost Index (CEPCI)}.
\newblock \textit{Chemical Engineering} [online], July 2021.
\newblock Available at: \url{https://www.chemengonline.com} (accessed: January
  31, 2023).

\bibitem{Clark:2018}
R.~L. Clark, L.~L. McGinley, H.~M. Purdy, T.~C. Korosh, J.~L. Reed, T.~W. Root,
  and B.~F. Pfleger.
\newblock Light-optimized growth of cyanobacterial cultures: growth phases and
  productivity of biomass and secreted molecules in light-limited batch growth.
\newblock {\em Metabolic engineering}, 47:230--242, 2018.

\bibitem{clippinger:2019}
J.~N. Clippinger and R.~E. Davis.
\newblock Techno-economic analysis for the production of algal biomass via
  closed photobioreactors: {Future} cost potential evaluated across a range of
  cultivation system designs.
\newblock Technical report, {National Renewable Energy Laboratory}, 2019.

\bibitem{biosteam}
Y.~Cortes-Peña, D.~Kumar, V.~Singh, and J.~S. Guest.
\newblock {BioSTEAM}: {A} fast and flexible platform for the design,
  simulation, and techno-economic analysis of biorefineries under uncertainty.
\newblock {\em ACS Sustainable Chemistry \& Engineering}, 8(8):3302--3310,
  2020.

\bibitem{dimpl:2010}
E.~Dimpl.
\newblock Small-scale electricity generation from biomass: {Experience} with
  small-scale technologies for basic energy supply.
\newblock Technical report, {German Federal Ministry for Economic Cooperation
  and Development (BMZ)}, 08 2010.

\bibitem{gebreslassie:2013}
B.~H. Gebreslassie, R.~Waymire, and F.~You.
\newblock Global optimization for sustainable design and synthesis of algae
  processing network for co2 mitigation and biofuel production using life cycle
  optimization.
\newblock {\em AIChE Journal}, 59(5):1599--1621, 2013.

\bibitem{hu:2022}
Y.~Hu, H.~Aguirre-Villegas, R.~A. Larson, and V.~M. Zavala.
\newblock Managing conflicting economic and environmental metrics in livestock
  manure management.
\newblock {\em ACS ES\&T Engineering}, 2(5):819--830, 2022.

\bibitem{larson:2016}
R.~A. Larson, M.~Sharara, L.~W. Good, P.~Porter, T.~Runge, V.~Zavala,
  A.~Sampat, and A.~Smith.
\newblock Evaluation of manure storage capital projects in the yahara river
  watershed.
\newblock Technical report, {University of Wisconsin-Extension, University of
  Wisconsin-Madison College of Agricultural and Life Sciences, Biological
  Systems Engineering}, 2016.

\bibitem{ma:2021}
J.~Ma, P.~Tominac, B.~F. Pfleger, and V.~M. Zavala.
\newblock Infrastructures for phosphorus recovery from livestock waste using
  cyanobacteria: {Transportation}, techno-economic, and policy implications.
\newblock {\em ACS Sustainable Chemistry \& Engineering}, 9(34):11416--11426,
  2021.

\bibitem{pipatmanomai:2009}
S.~Pipatmanomai, S.~Kaewluan, and T.~Vitidsant.
\newblock Economic assessment of biogas-to-electricity generation system with
  h2s removal by activated carbon in small pig farm.
\newblock {\em Applied Energy}, 86(5):669--674, 2009.

\bibitem{rogers:2014}
J.~N. Rogers, J.~N. Rosenberg, B.~J. Guzman, V.~H. Oh, L.~E. Mimbela,
  A.~Ghassemi, M.~J. Betenbaugh, G.~A. Oyler, and M.~D. Donohue.
\newblock A critical analysis of paddlewheel-driven raceway ponds for algal
  biofuel production at commercial scales.
\newblock {\em Algal Research}, 4:76--88, 2014.

\bibitem{solar:2020}
{Signature Solar}.
\newblock {EG4 BrightMount Solar Panel Ground Mount Rack Kit, 4 Panel Ground
  Mount, Adjustable Angle}.
\newblock Solar Signature Mounting Hardware [online], 2022.
\newblock Available at:
  \url{https://signaturesolar.com/eg4-brightmount-solar-panel-ground-mount-rack-kit-4-panel-ground-mount-adjustable-angle}
  (accessed January 20, 2023).

\bibitem{doe:2022}
{United States Department of Energy}.
\newblock {How Gas Turbine Power Plants Work}.
\newblock Office of Fossil Energy and Carbon Management [online], 2022.
\newblock Available at:
  \url{https://www.energy.gov/fecm/how-gas-turbine-power-plants-work}
  (accessed: October 6, 2022).

\bibitem{wang:2019}
H.~Wang, H.~A. Aguirre-Villegas, R.~A. Larson, and A.~Alkan-Ozkaynak.
\newblock Physical properties of dairy manure pre- and post-anaerobic
  digestion.
\newblock {\em Applied Sciences}, 9(13):2703, 2019.

\bibitem{Rasmussen:1996}
C.~K.~I. Williams and C.~E. Rasmussen.
\newblock Gaussian processes for regression.
\newblock In {\em Advances in Neural Information Processing Systems}, volume~8,
  pages 514--520. MIT Press, 1996.

\end{thebibliography}


\end{document}


\title{{\bf Supplementary Information}\\ On the Implementation of a Bayesian Optimization Framework for Interconnected Systems}

\author{Leonardo D. Gonz\'alez and Victor M. Zavala\thanks{Corresponding Author: victor.zavala@wisc.edu.}\\
  {\small Department of Chemical and Biological Engineering}\\
 {\small \;University of Wisconsin-Madison, 1415 Engineering Dr, Madison, WI 53706, USA}}

 \date{}
\maketitle

\renewcommand{\thefigure}{SI~\arabic{figure}}
\setcounter{figure}{0}
\renewcommand{\thetable}{SI~\arabic{table}}
\setcounter{table}{0}
\renewcommand{\thesection}{SI~\arabic{section}}
\setcounter{section}{0}

\noindent This document contains 16 pages, 2 figures, 6 tables, and includes:\\
\vspace{0.1in}

\begin{itemize}
\item \textbf{Figure \ref{si_fig:system_diag}:} Schematic diagram of the chemical process modeled in the first case study.
\item \textbf{Figure \ref{si_fig:renual_flow}:} Flow diagram of the biofertilizer production process.
\item \textbf{Table \ref{si_tab:thermo_constants}:} Relevant thermodynamic properties for the chemical process model. 
\item \textbf{Table \ref{si_tab:model_constants}:} Relevant parameters for the chemical process model. 
\item \textbf{Table \ref{si_tab:gen_renual}:} General biofertilizer production process information. 
\item \textbf{Table \ref{si_tab:yields_renual}:} Product yield factors. 
\item \textbf{Table \ref{si_tab:ISBL}:} Capital costs of process units. 
\item \textbf{Table \ref{si_tab:capital_renual}:} Variable operating costs of process units.
\end{itemize}

\newpage
\section{System Models}\label{ap:model_descriptions}

\subsection{Chemical Process Model}\label{ap:rxtr_flsh}

\begin{figure}[!htp]
	\begin{center}
	    \begin{subfigure}
	        \centering
	        \includegraphics[width=6in]{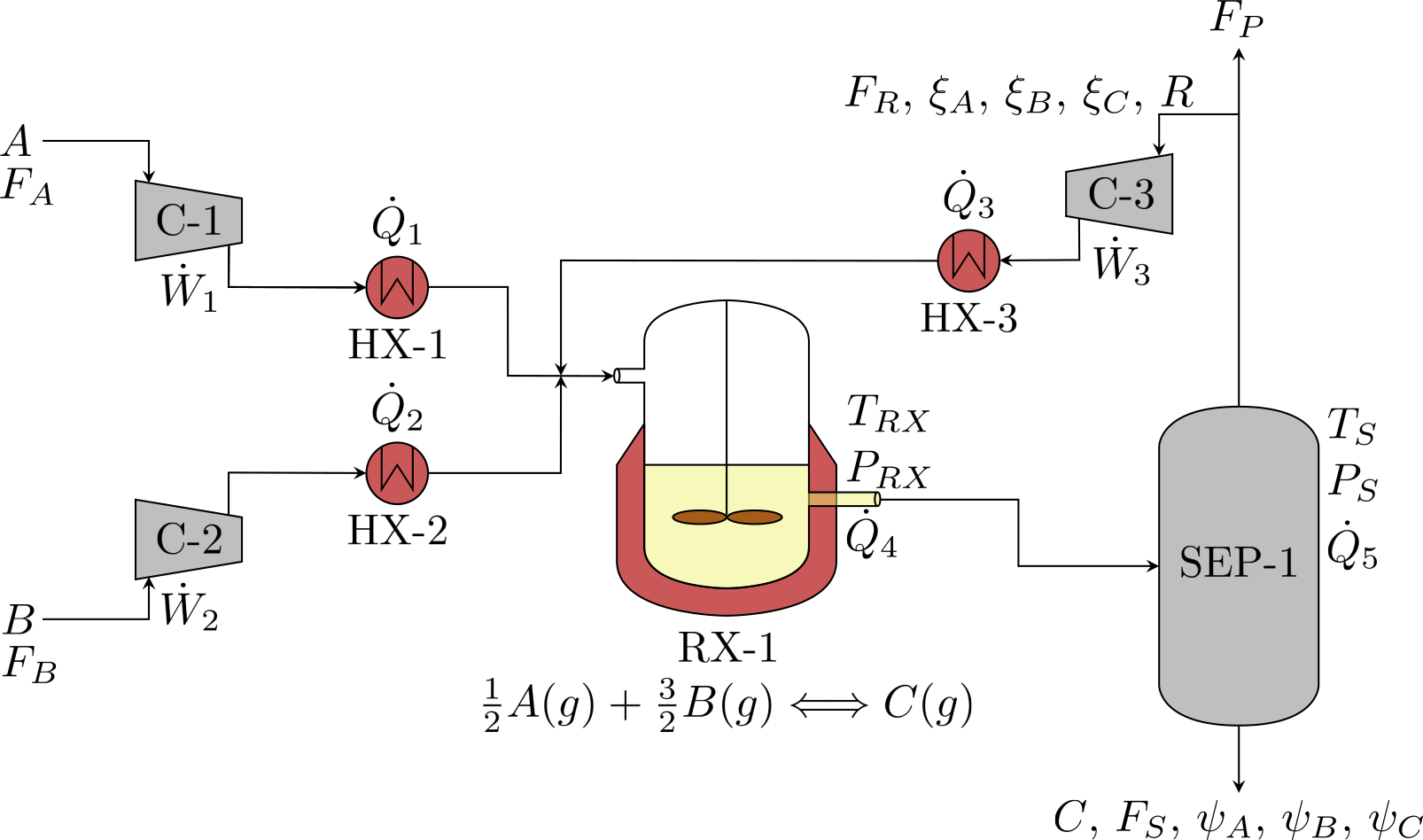}
	    \end{subfigure}
	    \caption{Schematic diagram of the chemical process modeled in the first case study.}
	    \label{si_fig:system_diag}
	\end{center}
\end{figure}

The chemical process model considers the generation of product $C$ from two reagents $A$ and $B$. As seen in Figure \ref{si_fig:system_diag}, these reagents are fed into the process where they are individually compressed and heated. The compressors are modeled with the {\tt IsentropicCompressor} module from the BioSTEAM library \cite{biosteam} using the properties of N$_2$, H$_2$, and NH$_3$ for $A$, $B$, and $C$ respectively and the {\tt IdealAcitivityCoefficients} and {\tt IdealFugacityCoefficients} methods. The models calculate the temperatures of the compressor outlet as well as the power load of the unit. The heaters are assumed to operate at constant pressure, and we can calculate their duties as:
\begin{subequations}\label{eq:hx_duty}
	\begin{align}
		\dot{Q} &=F(H_\textrm{out}-H_\textrm{in})\\
		&=F(H^{\circ}+R\cdot\textrm{ICPH}(T_\textrm{out})-H^{\circ}-R\cdot\textrm{ICPH}(T_\textrm{in}))\\
		&=FR\left(\textrm{ICPH}(T_\textrm{out})-\textrm{ICPH}(T_\textrm{in})\right)
	\end{align}
\end{subequations}
where $F$ is the material flowrate through the heater, $T_\textrm{in}$ and $T_\textrm{out}$ are the temperatures of the heater inlet and outlet respectively, $H^{\circ}$ is the standard enthalpy of the streams, and $R$ is the universal gas constant. The \textrm{ICPH} function measures the effect of temperature on enthalpy and is defined as: 
\begin{subequations}\label{eq:ICPH}
	\begin{align}
		\textrm{ICPH}(T) &= \Delta \alpha\left(T-T^{\circ}\right)+\frac{\Delta \beta}{2}\left(T^2-\left(T^{\circ}\right)^2\right) \notag \\ &+\frac{\Delta \gamma}{3}\left(T^3-\left(T^{\circ}\right)^3\right)-\Delta \zeta\left(T^{-1}-\left(T^{\circ}\right)^{-1}\right)
	\end{align}
\end{subequations}
where $T^{\circ}$ is the standard temperature (298 K) and 
\begin{subequations}\label{eq:ICPH_coefficients_simp}
	\begin{align}
		\Delta \alpha &= \sum_{i\in\{A,B,C\}} x_i \alpha_i\\
		\Delta \beta &= \sum_{i\in\{A,B,C\}} x_i \beta_i\\
		\Delta \gamma &= \sum_{i\in\{A,B,C\}} x_i \gamma_i\\
		\Delta \zeta &= \sum_{i\in\{A,B,C\}} x_i \zeta_i
	\end{align}
\end{subequations}
Here, $x_i$ is the molar fraction of species $i$ in the stream and $\alpha_i$, $\beta_i$, $\gamma_i$, and $\zeta_i$ are a set of molecule-specific coefficients.
\\

The hot streams are fed to a reactor, RX-1, where $A$ and $B$ undergo the following gas-phase reaction:
\begin{subequations}\label{rxn_net}
    \begin{gather}
        \frac{1}{2}A(g)+\frac{3}{2}B(g)\Longleftrightarrow C(g)
    \end{gather}
\end{subequations}
The reaction mixture is assumed to be an ideal gas, and, at a given temperature and pressure, the composition of the reactor effluent is calculated from the following:
\begin{equation}\label{eq:compositions}
	\prod_{i\in\{A,B,C\}} \left(y_i\right)^{\nu_i}=\prod_{i \in\{A,B,C\}} \left(\frac{n_{i0}+\nu_i\varepsilon}{n_0+\nu\varepsilon}\right)^{\nu_i}=\left(\frac{P_{RX}}{P^{\circ}}\right)^{-\nu}K
\end{equation}
where $y_i$ is the mol fraction of species $i$ in the reactor outlet and $\nu_i$ is its stoichiometric coefficient; $n_{i0}$ are the moles of $i$ in the feed stream, $\nu=\sum_{i}\nu_{i}$ and $n_{0}=\sum_{i}n_{i0}$; the extent of the reaction is denoted by $\varepsilon$; $P_{RX}$ is the pressure of the reactor and $P^{\circ}$ is standard pressure (1 bar). The equilibrium constant, $K$, is calculated as:
\begin{equation}\label{eq:equil}
	\log{K}=-\frac{\Delta G_{\textrm{rxn}}}{RT_{RX}}
\end{equation}
where $T_{RX}$ is the reactor temperature. The change in the Gibbs free energy of the reaction, $\Delta G_{\textrm{rxn}}$, is a function of temperature and has the form:
\begin{equation}\label{eq:deltaG}
	\Delta G_{\textrm{rxn}}=\Delta H^{\circ}_{\textrm{rxn}}-\frac{T_{RX}}{T^{\circ}}\left(\Delta H_{\textrm{rxn}}^{\circ}-\Delta G_{\textrm{rxn}}^{\circ}\right)+R\left(\textrm{ICPH}(T_{RX})-T_{RX}\cdot\textrm{ICPS}(T_{RX})\right)
\end{equation}
where $\Delta G^{\circ}_{\textrm{rxn}}$ and $\Delta H^{\circ}_{\textrm{rxn}}$ are the changes in Gibbs free energy and enthalpy of the reaction measured at standard temperature. Similarly to the $\textrm{ICPH}$, the $\textrm{ICPS}$ function measures the change in entropy due to temperature effects and is defined as:
\begin{subequations}\label{eq:ICPS}
	\begin{align}
		\textrm{ICPS}(T) &= \Delta \alpha\left(\log(T)-\log(T^{\circ})\right)+\Delta \beta\left(T-T^{\circ}\right) \notag \\ &+\frac{\Delta \gamma}{2}\left(T^2-\left(T^{\circ}\right)^2\right)-\Delta \zeta\left(T^{-2}-\left(T^{\circ}\right)^{-2}\right)
	\end{align}
\end{subequations}
Due to the reaction, the ICPH and ICPS coefficients are now defined as:
\begin{subequations}\label{eq:ICPH_coefficients_simp}
	\begin{align}
		\Delta \alpha &= \sum_{i\in\{A,B,C\}} \nu_i \alpha_i\\
		\Delta \beta &= \sum_{i\in\{A,B,C\}} \nu_i \beta_i\\
		\Delta \gamma &= \sum_{i\in\{A,B,C\}} \nu_i \gamma_i\\
		\Delta \zeta &= \sum_{i\in\{A,B,C\}} \nu_i \zeta_i
	\end{align}
\end{subequations}
Because the reaction occurs at constant temperature and pressure, the duty of the reactor, $\dot{Q}_4$ is calculated as:
\begin{equation}\label{eq:rxt_duty}
	\dot{Q}_4 = \dot{r}_C\left(\Delta H^{\circ}_{\textrm{rxn}}+R\cdot\textrm{ICPH}(T_{RX})\right)
\end{equation}
where $\dot{r}_C$ is the generation rate of $C$ in the reactor.
\\

The product stream that exits the reactor is fed to a separator, SEP-1, where it is cooled to $T_S$ at pressure $P_S$ to allow for the condensation and recovery of $C$. This unit is modeled with the {\tt Flash} module from the BioSTEAM library using the same properties and methods as the compressor models. From this model we obtain the flowrates and compositions of the resultant product and vapor streams as well as the energy requirements of the separator. A fraction, $1-R$, of the vapor stream that exits SEP-1 is purged and leaves the process. The remainder is re-compressed and heated before being fed back to RX-1.
\\

The performance of the system is expressed as a cost function (negative profit) that measures the consumption of the reagents along with production rate of $C$ and the corresponding utility requirements of the process at various recycle fractions and reactor and separator settings; it is formulated as:
\begin{subequations}\label{eq:ap_objective_rxtr_flsh}
    \begin{align}
	f_1(x, y(x)) & = \sum_{j\in\{A,B\}}w_{j0}F_{j}+F_S\sum_{i\in\{A,B,C\}}w_{i}\psi_i+w_3\left(\frac{\psi_C F_S-\bar{F}}{\bar{F}}\right)^2\\[5pt]
	f_2(x,  y(x)) & = \sum_{h=1}^5 w_h\dot{Q}_h, +w_e\sum_{k=1}^3\dot{W}_k \\[5pt]
	f(x, y(x)) & = f_1(x, y(x))+f_2(x, y(x))
    \end{align}
\end{subequations} 
where the total cost $f(x, y(x))$ is distributed among two functions $f_1$ and $f_2$. Here, $f_1$ considers the materials flowing in an out of the system: the first term represents the costs of $A$ and $B$ with $w_{j0}$ and $F_j$ denoting the cost and feed rate of each reagent; the second term measures the value of the product stream, with $F_S$ denoting its flow and $\psi_i$ and $w_{i}$ representing the fraction of species $i$ present in this stream and its corresponding unit value. The final term in $f_1$ is a quadratic penalty term meant to enforce a demand cap for the production of $C$, which is denoted by $\bar{F}$. The flow of energy through the system, and thereby the cost of utilities, is measured by $f_2$. The heating and cooling requirements of the feed and recycle heaters, RX-1, and SEP-1 are denoted by $\dot{Q}_h$, and these have a unit cost of $w_h$. Similarly, compressor $k$ has a power load of $\dot{W}_k$, and this is supplied at a unit cost of $w_e$.

\begin{table}[htp!]
\centering
\caption{Relevant thermodynamic \\ properties for the chemical process model.}
\label{si_tab:thermo_constants}
\resizebox{0.5\textwidth}{!}{
	\begin{tabular}{lrll}
		\toprule[1.5pt]
		\textbf{Parameter} & \textbf{Value} & \textbf{Units} & \textbf{Notes} \\
		$R$ & 8.314 & J/mol$\cdot$K & Universal gas constant \\
		$T^{\circ}$ & 298 & K & Standard temperature\\
		$P^{\circ}$ & 1.0 & bar & Standard pressure \\
		$\Delta H^{\circ}_{\textrm{rxn}}$ & 39200 & $\textrm{J/mol}$ & Standard heat of reaction \\
		$\Delta G^{\circ}_{\textrm{rxn}}$ & 32900 & $\textrm{J/mol}$ & Change in free energy of reaction at $T^{\circ}$ \\
		$\nu_A$ & $-1/2$ & --- & Stoichiometric coefficient of $A$ \\
		$\nu_B$ & $-3/2$ & --- & Stoichiometric coefficient of $B$ \\
		$\nu_C$ & $1$ & --- & Stoichiometric coefficient of $C$ \\
		$\alpha_A$ & 3.280 & --- & $\alpha$ coefficient of $A$ \\
		$\alpha_B$ & 3.249 & --- & $\alpha$ coefficient of $B$ \\
		$\alpha_C$ & 5.578 & --- & $\alpha$ coefficient of $C$ \\
		$\beta_A$ & \num{0.593e-3} & --- & $\beta$ coefficient of $A$ \\
		$\beta_B$ & \num{0.422e-3} & --- & $\beta$ coefficient of $B$ \\
		$\beta_C$ & \num{3.020e-3} & --- & $\beta$ coefficient of $C$ \\
		$\gamma_A$ & $0.000$ & --- & $\gamma$ coefficient of $A$ \\
		$\gamma_B$ & $0.000$ & --- & $\gamma$ coefficient of $B$ \\
		$\gamma_C$ & $0.000$ & --- & $\gamma$ coefficient of $C$ \\
		$\zeta_A$ & \num{0.040e5} & --- & $\zeta$ coefficient of $A$ \\
		$\zeta_B$ & \num{0.083e5} & --- & $\zeta$ coefficient of $B$ \\
		$\zeta_C$ & \num{-0.186e5} & --- & $\zeta$ coefficient of $C$ \\
		\bottomrule[1.5pt] 
	\end{tabular}
}
\end{table}

\begin{table}[htp!]
\centering
\caption{Relevant parameters \\ for the chemical process model.}
\label{si_tab:model_constants}
\resizebox{0.5\textwidth}{!}{
	\begin{tabular}{lrll}
		\toprule[1.5pt]
		\textbf{Parameter} & \textbf{Value} & \textbf{Units} & \textbf{Notes} \\
		$F_A$ & 1000 & kmol/hr & Feed flow of $A$ \\
		$F_B$ & 3000 & kmol/hr & Feed flow of $B$ \\
		$\bar{F}$ & 1900 & kmol/hr & Target production rate of $C$ \\
		$w_{A0}$ & \num{6.00} & USD/kmol & Cost of $A$ \\
		$w_{B0}$ & \num{1.40} & USD/kmol & Cost of $B$ \\
		$w_{A}$ & \num{0.00} & USD/kmol & Value of $A$ in product stream \\
		$w_{B}$ & \num{0.00} & USD/kmol & Value of $B$ in product stream\\
		$w_{C}$ & \num{8.50} & USD/kmol & Value of $C$ in product stream\\
		$w_{h}$ & \num{1.92e-2} & USD/MJ & Cost of heating utility \\
		$w_{h}$ & \num{5.00e-3} & USD/MJ & Cost of cooling utility \\
		$w_{e}$ & \num{1.42e-1} & USD/kWh & Cost of power utility \\
		\bottomrule[1.5pt] 
	\end{tabular}
}
\end{table}

\newpage
\subsection{Biofertilizer Production Process Model}\label{ap:pbr_model}

\begin{figure}[!htp]
	\begin{center}
	    \begin{subfigure}
	        \centering
	        \includegraphics[width=0.9\textwidth]{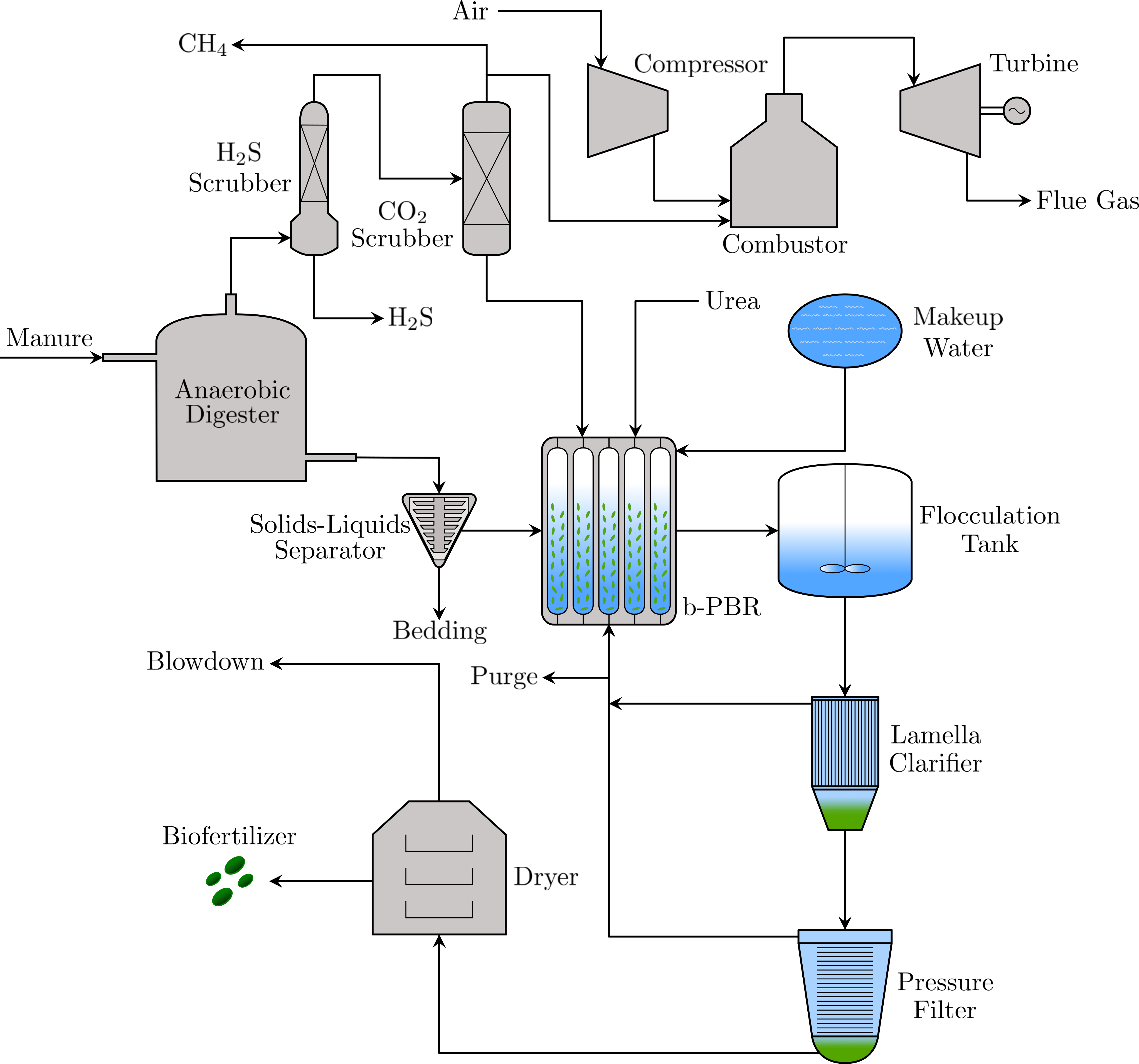}
	    \end{subfigure}
	    \caption{Flow diagram of the biofertilizer production process.}
	    \label{si_fig:renual_flow}
	\end{center}
\end{figure}

The biofertilizer production facility we consider, shown in Figure \ref{si_fig:renual_flow}, considers the recovery of 10.59 tonnes/yr of phosphorus (P) from the 20830 tonnes/yr of manure generated at a hypothetical 1000 animal unit dairy farm. The manure is initially fed into an anaerobic digester to generate biogas. The biogas, a mixture largely composed of CH$_4$, CO$_2$, and H$_2$S, is sent to a pair of scrubbers that remove the carbon dioxide and hydrogen sulfide. The resulting product is a high purity methane stream that can either be exported or burned on-site to generate electricity. 
\\

The digested manure (also referred to as digestate) is passed through a solids-liquids separator (SLS) that removes the suspended solids present in the digestate. The produced extrudate is then fed to a series of bag photobioreactors (b-PBRs), along with any urea required to provide additional nitrogen (N), where it is used as a growth medium for cyanobacteria (CB) cultivation. The CB growth is assumed to be light-limited and is simulated as described by \cite{Clark:2018}:
\begin{subequations}
	\begin{gather}
		X(t) =X_0\exp(-\kappa t)+X_s\left(1-\exp(-\kappa t)\right) \\
		X_S = \frac{\eta\frac{I_0S}{V}}{m_\nu}\\
		\kappa = Y_{X\nu}m_\nu
	\end{gather}
\end{subequations} 
Here $X$ is the CB concentration at any given time, $X_0$ is the initial CB concentration, and $X_S$ is the steady-state $(t=\infty)$ CB concentration; the biomass yield per photon is denoted by $Y_{X\nu}$, and $m_\nu$ is the light energy required for cell maintenance; $\eta$ is the photosynthetic efficiency of the cell (i.e., how much of the energy in each photon is captured and converted into biomass); $S$ and $V$ are the irradiated reactor surface area and total reactor volume respectively, and $I_0$ is the incident light intensity at the surface of the reactor.  
\\

Once the CB culture is ready to be harvested, it is sent to a flocculation tank where self-flocculation is induced. The formed flocs are then fed into a lamella clarifier where they are allowed to settle out of solution. The CB sludge that forms at the bottom of the clarifier is collected and passed through a pressure filter that produces a concentrated cyanobacteria solution. The solution is transferred to a thermal dryer where the moisture removal process is completed, yielding a dry CB biomass product. The CB-free water that exits the lamella clarifier and pressure filter is mixed, and a fraction of it is recycled back to the reactors, while the remainder is purged to prevent the buildup of impurities within the system. Depending on the flowrate of the purge, fresh make-up water may need to be added to the reactors if the water in the incoming manure is not enough to account for this loss. 
\\

The CB production rate is set based on the phosphorus density of the bacteria cells, $\rho_P$ (g P/g CB). Given an incoming P flowrate, $m_P$, the CB needed to fully incorporate the nutrient load is calculated as:
\begin{equation}
	m_{CB} = \frac{m_P}{\rho_P}
\end{equation}
From this, we can determine the amount of urea, $m_U$, required to meet the N demands of the cells based on a specified nitrogen density, $\rho_N$  (g N/g CB):
\begin{equation}
	m_{U} = \max\left\{0, x_{UN}\left(\rho_Nm_{CB}-m_N\right)\right\}
\end{equation}
where $m_N$ is the flowrate of $N$ in the reactor feed, and $x_{UN}$ is the mass fraction of nitrogen in urea. The amount of fresh makeup water required is also calculated based on the CB production rate and the final titer of the CB culture:
\begin{equation}
	m_{FW} = \max\left\{0, m_{CB}\left(\frac{\rho_W}{X(t_b)}-1\right)-(m_{RW}+m_{W})\right\}
\end{equation}
where $t_b$ is the reactor batch time, $\rho_W$ is the density of water, $m_{RW}$ is the flowrate of the recycled water stream, and $m_W$ is the flowrate of water supplied to the b-PBRs from the manure.
\\

The minimum selling price (MSP) of the biofertilizer is determined by calculating the price, $p_{CB}$, that it must be sold at to for the process to achieve a specified discounted return on investment (DROI) target (15\% in this study). The DROI is a profitability measure that is defined as the discount rate, $i$, that results in a net present value (NPV) of zero at the end of a process' lifetime. We define the NPV as:
\begin{equation}\label{eq:NPV}
	\textrm{NPV}=C+\sum_{j=1}^{T}P(1+i)^{-j}
\end{equation}
where $C$ is the total capital investment (TCI) of the of the process, $P$ is the annual after-tax profit (AATP) and $T$ is the process lifetime (in years).
\\

The TCI is based on the installed costs of the required units and is calculated according to their size. If sizing and costing correlations are available, these are used to calculate these values. Otherwise, they can be obtained from price data found in the literature for similar units using capacity correlations and price indices:
\begin{equation}\label{eq:installed_cost}
	c_k=c_k^{\prime}\left(\frac{S_k}{S^{\prime}_k}\right)^{\phi_k}\left(\frac{PI}{PI^{\prime}_k}\right)
\end{equation}
where $c_k$, $S_k$, and $\phi_k$ are cost, size, and scaling factor respectively of the $k^{\textrm{th}}$ unit; $c_k^{\prime}$ and $S^{\prime}_k$ are the cost and size of a reference unit; $PI$ is the value of the price index for the year 2020 and $PI^{\prime}_k$ is the price index for the year in which the quote for the reference unit was made. We can then define the inside battery limit (ISBL) cost of the process as:
\begin{equation}\label{eq:installed_cost}
	c_{IS}=\sum_{k}^{K}c_k
\end{equation}
which serves as the basis quantity from which the TCI can be calculated:
\begin{subequations}
	\begin{align}
		c_{OS} &= 0.4c_{IS}\\
		c_{ENG} &= 0.3(c_{IS}+c_{OS})\\
		c_{CON} &= 0.2(c_{IS}+c_{OS}).
	\end{align}
\end{subequations}
where $c_{OS}$, $c_{ENG}$, $c_{CON}$ are the outside battery limits (OSBL), engineering, and contingency costs associated with constructing the process and bringing it online. The total capital investment required is then:
\begin{equation}
	C = c_{IS}+c_{OS}+c_{ENG}+c_{CON}
\end{equation}
\\

The AATP is the net income generated by the process and is expressed as:
 \begin{equation}
 	P = (1-r)\left(p_{CB}m_{CB}+p_{NG}m_{NG}+p_{EL}\dot{w}-O-d\right)+d
 \end{equation}
 where $m_{CB}$, $m_{NG}$, and $\dot{w}$ are the production rates of cyanobacteria, methane, and electricity respectively, and these are sold at unit price of $p_{CB}$, $p_{NG}$, and $p_{EL}$. The tax rate levied on the process revenue is denoted by $r$ (21\% in the US); $d$ is annual depreciation of the purchased capital equipment and is calculated using a straight-line depreciation scheme assuming a salvage value of 0:
\begin{equation}
 	d = \frac{c_{IS}}{T}
 \end{equation}
The total operating cost (TOC) of the process, $O$, is composed of two parts, fixed operating costs (FOCs) and variable operating costs (VOCs). FOCs include annual costs that are not tied to production levels and must be paid in full every year the plant operates. This includes expenses such as maintenance, operations fees such as licensing costs and property taxes or leasing costs, overhead costs like insurance, and labor. Maintenance, operations, and overhead costs are calculated as a fraction of the ISBL:
\begin{subequations}
	\begin{align}
		c_{MT} &= 0.05c_{IS}\\
		c_{OP} &= 0.025c_{IS}\\
		c_{OV} &= 0.05c_{IS}\\
	\end{align}
\end{subequations}
Labor costs are calculated as a function of the size of the reaction section, $SA$ (in acres):
\begin{equation}\label{eq:labor}
 	c_{LB} = c_{LB}^{\prime}\left(\frac{SA}{SA^{\prime}}\right)
 \end{equation}
where $c_{LB}^{\prime}$ is the labor cost at a reference facility of size $SA^{\prime}$
\\

Variable operating costs consists of items like utilities, raw materials, and storage and transportation that fluctuate with production levels. As we do not consider distribution costs, and we assume that manure and any make-up water required are provided free of charge, the VOCs in this study consist of the utility, urea, and bag replacement costs. The utility requirements are obtained either from correlations or by scaling reported values for similar units from the literature to the desired size, similar to \eqref{eq:labor}. We include a scale factor on the bag replacement and reactor mixing costs of the form:
\begin{equation}\label{eq:SA_to_V}
 	\lambda_{SV}\left(\frac{S}{V}\right) = \max\left\{1, 3\left(\frac{\frac{S}{V}}{\left(\frac{S}{V}\right)_0}\right)-2\right\}
 \end{equation}
where $\left(\frac{S}{V}\right)_0$ is the surface area to volume ratio of the base design. This penalty reflects the fact that higher $\frac{S}{V}$ values require more complex geometries which increase the manufacturing costs of the bags and also increase the flow turbulence within the reactors, resulting in a higher mixing energy requirement. A penalty is also added to the labor cost:
\begin{equation}\label{eq:labor_penalty}
 	\lambda_{LB}(t_b) = \max\left\{0, 0.05(t_{b0}-t_b)\right\}
 \end{equation}
where $t_{b0}$ is the batch time of the base design and $\lambda_{LB}(t_b) $ has units of MMUSD/yr. This reflects the fact that smaller values of $t_b$ increase frequency at which the b-PBRs must be emptied, cleaned, and refilled and by extension the labor intensity of the process. Finally, we scale the value of $X_S$ by a factor of 0.32, as the model growth model used likely over-predicts the true titer that is achievable. This places the value of $X_S$ within the range of 1-2 g/L for the range of $t_b$ we consider, which is the range of titers commonly seen in practice at commercial CB cultivation facilities.

\newpage
\begin{table}[htp!]
\centering
\caption{General biofertilizer production process information \cite{larson:2016, wang:2019, cepci:2020, Clark:2018}.}
\label{si_tab:gen_renual}
\resizebox{0.45\textwidth}{!}{
	\begin{tabular}{lrll}
		\toprule[1.5pt]
		\textbf{Parameter} & \textbf{Value} & \textbf{Units} & \textbf{Notes} \\
		$m_M$ & 20830 & tonnes/yr & Manure feed flowrate\\
		$m_P$ & 9.64  & tonnes/yr & P in SLS extrudate\\
		$m_N$ & 10.60 & tonnes/yr & N in SLS extrudate\\
		$m_W$ & 18030 & tonnes/yr & Water in SLS extrudate\\
		$Y_{X\nu}$ & \num{2.02E-9} & kg/\textmu mol & Biomass yield per photon\\
		$m_{\nu}$ & 255 & \textmu mol/kg$\cdot$s & CB cell maintenance light needs\\
		$\eta$ & 0.24 & --- & Photosynthetic efficiency of CB\\
		$X_0$ & 0.03 & g/L & Initial CB concentration\\		
		$I_0$ & 350 & \textmu mol/m$^{2}\cdot$s & Light intensity\\
		$\left(\frac{S}{V}\right)_0$ & 15.4 & m$^{-1}$ & $S:V$ ratio of base design\\
		$\sigma$ & 70 & kg/m$^{2}$ & Areal density of b-PBRs\\
		$t_{b0}$ & 30 & days & Batch time of base design\\
		$\rho_{P0}$ & 0.023 & g P/ g CB & P density of base design\\
		$\rho_N$ & 0.05 & g N/g CB & N denisty of CB\\
		$x_{UN}$ & 0.467 & --- & Mass fraction of nitrogen in urea\\ 
		$PI$ & 596.2 & --- & Cost index for 2020\\
		$T$ & 10 & years & Lifetime of ReNuAl process \\
		$p_{EL}$ & 0.11 & USD/kW-hr & Electricity price \\
		$p_{NG}$ & 5.84 & USD/1000 SCF & Natural gas price \\
		$\rho_{W}$ & 1000 &  kg/m$^3$ & Water density \\
		$\rho_{BG}$ & 1.2 &  kg/m$^3$ & Biogas density \\
		$\rho_{NG}$ & 0.72 & kg/m$^3$ & Natural gas density\\
		\bottomrule[1.5pt] 
	\end{tabular}
}
\end{table}

\begin{table}[htp!]
\centering
\caption{Product yield factors.}
\label{si_tab:yields_renual}
\resizebox{0.85\textwidth}{!}{
	\begin{tabular}{llrl}
		\toprule[1.5pt]
		\textbf{Product} & \textbf{Unit Operation} & \textbf{Yield} & \textbf{Notes} \\
		CH$_4$\cite{dimpl:2010} & Anaerobic Digester & \num{3.09E-2} kg/kg manure & CH$_4$ from biogas production \\
		CO$_2$\cite{dimpl:2010} & Anaerobic Digester & \num{1.66E-2} kg/kg manure & CO$_2$ from biogas production \\
		H$_2$S\cite{dimpl:2010} & Anaerobic Digester & \num{1.14E-4} kg/kg manure & H$_2$S from biogas production \\
		Electricity\cite{doe:2022} & Electricity Generator &  4.33 kW-hr/kg CH$_4$ & assumes gas turbine efficiency of 0.3 and CH$_4$ energy content of 1000 BTU/scf \\
		Bedding\cite{villegas:2019} & Solid-Liquid Separator & 0.09 kg/kg manure & P and N are assumed to be uniformly distributed in digestate liquid and solid fractions  \\
		Primary Dewatering Product\cite{rogers:2014} & Lamella Clarifier & \num{1.60E-2} kg/L & $-$ \\ 
		Secondary Dewatering Product\cite{ma:2021} & Pressure Filter & 0.27 kg/L & $-$ \\
		\bottomrule[1.5pt] 
	\end{tabular}
}
\end{table}

\begin{table}[htp!]
\centering
\caption{Capital costs of process units.}
\label{si_tab:ISBL}
\resizebox{0.85\textwidth}{!}{
	\begin{tabular}{lrlrrrl}
		\toprule[1.5pt]
		\textbf{Item} & $\boldsymbol{c_k^{\prime}}$  & \textbf{Units} & $\boldsymbol{PI_k^{\prime}}$ & \textbf{Size Ratio} & $\boldsymbol{\phi_k}$ & \textbf{Notes} \\
		Anaerobic Digester (AD)\cite{hu:2022} & $937.1\left(m_{in}\right)^{0.6}+75355$ & USD & 539.1 & $-$ & $-$ & $m_{in}$ denotes unit capacity (tonne/yr) \\
		Solid-Liquid Separator (SLS)\cite{hu:2022} & $14.9m_{in}+1786.9\log\left(m_{in}\right)-9506.6$ & USD & 556.7 & $-$ & $-$ & $m_{in}$ denotes unit capacity (lb/hr) \\
		Electricity Generator\cite{hu:2022} & $0.67c_{AD}x_{CH4}$ & USD & 539.1 & $-$ & $-$ & \begin{tabular}[t]{@{}l@{}} $c_{AD}$ denotes cost of AD (USD) \\ $x_{CH4}$ denotes fraction of biogas produced used for electricity \end{tabular} \\
		H$_2$S Scrubber\cite{pipatmanomai:2009} & 348 & USD & 521.9 & \num{2.59E1} & 0.6 & $-$ \\
		CO$_2$ Scrubber\cite{gebreslassie:2013} & 13.1 & MMUSD & 444.2 & \num{4.37E-4} & 0.8 & $-$ \\
		Photobioreactors\cite{clippinger:2019, solar:2020} & 279 & USD & 556.8 & \num{1.08E5} & 0.6 & cost of structural system, bag costs are included in TOC \\
		Flocculation Tank\cite{rogers:2014} & 0.115 & MMUSD & 585.7 & \num{1.57E-3} & 0.6 & Scaling is in terms of CB mass \\
		Lamella Clarifier\cite{rogers:2014} & 2.50 & MMUSD & 585.7 & \num{1.57E-3} & 0.6 & Scaling is in terms of CB mass \\ 
		Pressure Filter\cite{ma:2021} & 0.137 & MMUSD & 381.8 & \num{2.39E-1} & 0.6 & $-$ \\
		Dryer\cite{ma:2021} & 0.706 & MMUSD & 539.1 & \num{2.20E3} & 0.6 & $-$ \\ 
		\bottomrule[1.5pt] 
	\end{tabular}
}
\end{table}

\begin{table}[htp!]
\centering
\caption{Variable operating costs of process units.}
\label{si_tab:capital_renual}
\resizebox{0.85\textwidth}{!}{
	\begin{tabular}{lrlll}
		\toprule[1.5pt]
		\textbf{Item} & \textbf{Operating Cost} & \textbf{Units} & \textbf{Utilities} & \textbf{Notes} \\
		Anaerobic Digester\cite{hu:2022} & $0.096c_{AD}$ & USD/yr & electricity & $c_{AD}$ denotes cost of AD (USD) \\
		Solid-Liquid Separator\cite{hu:2022} & $0.488m_{in}+0.1c_{SLS}$ & USD/yr & electricity & $m_{in}$ and $c_{SLS}$ denote capacity and cost of SLS (USD) \\
		H$_2$S Scrubber\cite{pipatmanomai:2009} & 66.7 & USD/tonne biogas & activated carbon &  gas removal via carbon bed adsorption \\
		CO$_2$ Scrubber\cite{gebreslassie:2013} & 40.0 & USD/tonne CO$_2$ & amine solution, steam & gas removal via amine scrubbing \\
		Photobioreactors\cite{clippinger:2019} & 12100 & USD/acre/yr & electricity, bags, urea, water & $-$ \\
		Flocculation Tank\cite{rogers:2014} & 100 & USD/tonne CB & electricity & source includes cost of chemical flocculant \\
		Lamella Clarifier\cite{rogers:2014} & 0.43 & USD/tonne CB & electricity & $-$ \\ 
		Pressure Filter\cite{ma:2021} & 2.06 & USD/tonne CB & electricity & $-$ \\
		Dryer\cite{ma:2021} & 19.3 & USD/tonne water & natural gas & basis is in terms of water removed \\
		\bottomrule[1.5pt] 
	\end{tabular}
}
\end{table}

\newpage
\section{Setup of Intermediate Function GP Models}\label{ap:GP_model_setup}

\subsection{Chemical Process Optimization Study}\label{ap:rxtr_flsh_intermediates}

In the composite function BO problem, we treat the reactor and separator are as black-boxes. The behavior of these units is approximated using a set of GP surrogate models, $\mathcal{GP}_y^{\ell}$, that estimate the following intermediate functions, $y(x)$: the duties of RX-1 and SEP-1, $\dot{Q}_4$ and $\dot{Q}_5$ respectively, the product to purge ratio of $A$, the feed to purge ratio of $B$, and purge to product ratio of $C$. We define these last three variables as:
\begin{subequations}\label{eq:ap_objective_rxtr_flsh}
    \begin{align}
	\eta_A &= \frac{\psi_A F_S}{\xi_A F_P} \\
	\eta_B &= \frac{\xi_B F_P}{F_B} \\
	\eta_C & = \frac{\xi_C F_P}{\psi_C F_S}
    \end{align}
\end{subequations} 
where $F_P$ is the flowrate of the purge and $\xi_i$ is the fraction of species $i$ in this stream. These intermediates have the following lower and upper feasibility bounds:
\begin{subequations}\label{eq:feas_bnds_rxt_flsh_rcyl}
    \begin{align}
	\hat{l}_y & = [10^{-6}, 10^{-6}, 10^{-6}, 10^{-6}, 10^{-6}]\\
	\hat{u}_y & = [1, \infty, \infty, \infty, \infty]
    \end{align}
\end{subequations} 
Because $B$ is essentially non-condensable and only present in trace amounts in the product stream, $1-\eta_B$ is can be assumed to be equal to the conversion of $B$ in the reactor. As a result, the generation rates of $A$, $B$, and $C$ can be calculated from this variable:
\begin{subequations}\label{eq:rxtr_generation}
    \begin{align}
	\dot{r}_B &= -(1-\eta_B)F_B \label{eq:rxtr_generation_B} \\
	\dot{r}_A &= \frac{\nu_A}{\nu_B}\dot{r}_B \\
	\dot{r}_C &= \frac{\nu_C}{\nu_B}\dot{r}_B
    \end{align}
\end{subequations}
Note that the negative sign in \eqref{eq:rxtr_generation_B} is due to the fact that B consumed by the reaction. Combining these flowrates with $\eta_A$ and $\eta_C$ we can calculate the flowrates and compositions of the remaining streams. The purge and recycle streams are specified by:
\begin{subequations}\label{eq:purge_info}
    \begin{align}
	&\xi_A F_P = \frac{F_A+\dot{r}_A}{1+\eta_A} \\
	&\xi_B F_P = \eta_B F_B\\
	&\xi_C F_P = \frac{\eta_C\dot{r}_C}{1+\eta_C}\\
	&F_P = \xi_A F_P+\xi_B F_P+\xi_C F_P\\
	&F_R = \frac{F_PR}{1-R}
    \end{align}
\end{subequations}
where $F_R$ is flowrate of the recycle stream; note that the purge and recycle streams have the same composition. The product stream can similarly be defined:
\begin{subequations}\label{eq:product_info}
    \begin{align}
	&\psi_A F_S = \frac{\eta_A(F_A+\dot{r}_A)}{1+\eta_A} \\
	&\psi_B F_S = 0\\
	&\psi_C F_S = \frac{\dot{r}_C}{1+\eta_C}\\
	&F_S = \psi_A F_S+\psi_B F_S+\psi_C F_S.
    \end{align}
\end{subequations}
Combining these expressions with the compressor and heater models (which are assumed to be white-boxes) and the GP estimates for $\dot{Q}_4$ and $\dot{Q}_5$, we can calculate all of the values in \eqref{eq:ap_objective_rxtr_flsh}.
\\

Due to the presence of the recycle stream, we initially surmised that $\eta_B$ should be a function of the five inputs as it is essentially a measure of the extent of the reaction. However, automatic relevance determination (ARD) \cite{Rasmussen:1996} showed that $P_S$ does not appear to have much of an effect on $\eta_B$. This is likely due to fact that $T_S$ is able to capture most of the variability from the separation process. As a result, the GP model of $\eta_B$ took only $T_{RX}$, $P_{RX}$, $R$, and $T_S$ as inputs. We then nested $\eta_B$ within the models of $\eta_A$ and $\eta_C$. These measure the performance of the separator, which is dependent on the composition of the separator feed in addition to $T_S$ and $P_S$. As the reactor outlet is the separator feed and is directly dependent on the value of $\eta_B$, $\eta_A$ and $\eta_C$ should exhibit a similar dependence. This allowed us to avoid having to explicitly consider the recycle fraction and reactor temperature and pressure (all of which affect the reactor outflow), thereby reducing the input dimension of the GP models for these intermediates from five to three. Using a similar approach, we determined that the reactor heat duty could effectively be modeled using $T_{RX}$, $P_{RX}$, $\eta_A$ and $\eta_B$ as the latter two variables are closely correlated to flowrate and composition of the recycle stream, while the former two establish the equilibrium point of the reactor. Similarly, the GP model of $\dot{Q}_5$ used $T_{RX}$, $R$, and $\eta_B$ along with $T_S$, and $P_S$ as these inputs are closely correlated to the flowrate and composition of the separator feed as well as the simple and latent heat released by the condensation of the product stream. Note that this approach highlights the high degree of customization composite function BO affords in the selection of GP model inputs. In addition to improving model quality, this can allow BO to be extended to high-dimensional systems where standard BO runs into scalability issues, as the individual intermediates do not have to be modeled with the full set of inputs. 

\subsection{Photobioreactor Design Study}\label{ap:pbr_intermediates}

Because of the novelty of the presented CB cultivation system, mature models for this unit operation are not yet available. The remaining systems (AD, biogas purification, and CB harvesting) are established technologies and models of these units can be found in the literature. As a result, we chose to treat these systems as white-boxes and only model the b-PBR as a black-box. The design variables of interest are the reactor surface area to volume ratio, $\frac{S}{V}$, batch time, $t_b$, and the phosphorus cell density of the CB, $\rho_P$. 
\\

The intermediate functions we chose to estimate are the required reactor volume $V$ and the CB titer at harvest time $X$; we define their feasibility bounds to be:
\begin{subequations}\label{eq:feas_bnds_efri}
    \begin{align}
	\hat{l}_y & = [10^{-6}, 10^{-6}]\\
	\hat{u}_y & = [10^6, 10]
    \end{align}
\end{subequations}
Using these variables, we can fully specify the outlet flow and size of the b-PBRs:
\begin{subequations}\label{eq:pbr_spec}
    \begin{align}
	&SA = \frac{\rho_WV}{\sigma}\\
	&m_{CB} = X\left(\frac{V}{t_b/365}\right)\\
	&m_{PBR} = \rho_W\left(\frac{V}{t_b/365}\right) 
    \end{align}
\end{subequations}
where $m_{PBR}$ is the mass flow rate out of the reactor. Note that we approximated the density of the culture as $\rho_W$ due to the relatively low concentration of the CB; the division of $t_b$ by 365 was done to convert this quantity from days to years. The GP models developed for $X$ and $V$ both took in the full set of inputs ($\frac{S}{V}$, $t_b$, and $\rho_P$). This was due to the fact that these variables have a significant impact on the throughput and size of the reactor. As a result, this case study demonstrated how composite function BO can be used to select sample points that aid in the development of a surrogate model for a unit of interest, while simultaneously driving the system the unit is a part of to an optimal configuration. This essentially allows us to complete two tasks at once and significantly reduces the probability of sampling at points in highly sub-optimal areas where the unit would likely never operate. As a result, the developed model is highly refined in the regions around the located optima and, therefore, is arguably more useful.

\clearpage
\bibliographystyle{abbrv}
\bibliography{./root}